\documentclass[journal]{IEEEtran}

\ifCLASSINFOpdf
\else
\fi

\usepackage{setspace}
\usepackage{duckuments} 
\usepackage{times}
\usepackage{subfig}
\usepackage{graphicx}
\usepackage{amsmath,amsthm}
\usepackage{amssymb,wasysym}
\usepackage{mathrsfs}
\usepackage{cite}
\usepackage{xcolor}
\usepackage{url}
\usepackage[lined,linesnumbered,ruled]{algorithm2e}
\usepackage{algorithmic}
\usepackage{wrapfig}
\usepackage{verbatim}
\usepackage{dsfont}
\usepackage{multirow}

\ifodd 0
\newcommand{\rev}[1]{{\color{green}#1}} 
\newcommand{\needrev}[1]{{\color{red}#1}} 
\newcommand{\waitrev}[1]{{\color{orange}#1}} 
\newcommand{\peerrev}[1]{{\color{pink}#1}} 
\newcommand{\jnlrev}[1]{{\color{purple}#1}} 
\else
\newcommand{\rev}[1]{#1}
\newcommand{\needrev}[1]{#1}

\newcommand{\waitrev}[1]{#1}
\newcommand{\peerrev}[1]{#1}
\newcommand{\jnlrev}[1]{#1}
\fi

\begin{document}

\newcommand{\name}{FedAlign-MoE\xspace}

\title{\jnlrev{Federated Mixture-of-Experts Alignment on Mobile Edge Networks under Data Heterogeneity}}

\author{Zihan Fang, Qianru Wang, Haonan An, Zheng Lin, Yiqin Deng,~\IEEEmembership{Member,~IEEE},\\Symeon Chatzinotas,~\IEEEmembership{Fellow,~IEEE}, and Yuguang Fang,~\IEEEmembership{Fellow,~IEEE}

\thanks{{The research work described in this paper was conducted in the JC STEM Lab of Smart City funded by The Hong Kong Jockey Club Charities Trust under Contract 2023-0108. This work was also supported in part by the Hong Kong SAR Government under the Global STEM Professorship and Research Talent Hub, and in part by the Hong Kong Innovation and Technology Commission under InnoHK Project CIMDA. The work of Qianru Wang was supported in part by the National Natural Science Foundation of China under Grant No. 62302367 and Hong Kong Scholar Program.
The work of Yiqin Deng was supported in part by the National Natural Science Foundation of China under Grant No. 62301300. 
}}

\thanks{Z. Fang, H. An, and Y. Fang are with Hong Kong JC STEM Lab of Smart City and Department of Computer Science, City University of Hong Kong, Kowloon, Hong Kong, China (e-mail: zihanfang3-c@my.cityu.edu.hk; haonanan2-c@my.cityu.edu.hk; my.fang@cityu.edu.hk).}
\thanks{Q. Wang is with the School of Computer Science and Technology, Xidian University, Xi'an, China, and  Hong Kong JC STEM Lab of Smart City and Department of Computer Science, City University of Hong Kong, Kowloon, Hong Kong, China (email: wangqianru@xidian.edu.cn).}
\thanks{Z. Lin and S. Chatzinotas are with the Interdisciplinary Centre for Security, Reliability and Trust (SnT), University of Luxembourg, L-1855 Luxembourg, Luxembourg (e-mail: zhenglin@ieee.org; Symeon.Chatzinotas@uni.lu).}
\thanks{Yiqin Deng is with the School of Data Science, Lingnan University, Tuen Mun, Hong Kong, China (email: yiqindeng@ln.edu.hk).}
}


\maketitle

\begin{abstract}
The growing demand for on-device large language model (LLM) services on mobile edge devices has driven the adoption of Mixture-of-Experts (MoE) architectures, which scale model capacity with limited computation. Since fine-tuning MoE-based LLMs relies on privacy-sensitive local data, federated learning (FL) offers a natural paradigm for collaborative training without exposing raw data. However, integrating MoE-based LLM fine-tuning into FL faces two critical challenges caused by data heterogeneity across clients: (i) divergent local data distributions drive clients to develop distinct gating preferences, so direct parameter aggregation yields a one-size-fits-none global gating network; and (ii) same-indexed experts develop disparate semantic roles across devices, leading to expert semantic blurring and degraded specialization. To address these challenges, we propose \name, a federated aggregation alignment framework for edge computing systems that jointly enforces routing consistency and expert semantic alignment. Specifically, \name aggregates gating behaviors by aligning routing distributions through consistency weighting and optimizes local gating networks through distribution regularization, maintaining cross-client stability while preserving discriminative local gating preferences. Meanwhile, \name quantifies the semantic consistency of same-indexed experts across devices and selectively aggregates semantically aligned experts, ensuring stable and specialized global experts. Extensive experiments demonstrate that \name outperforms state-of-the-art benchmarks, achieving faster convergence and higher accuracy in non-IID federated environments with lightweight computation and efficient communication.
\end{abstract}


\IEEEpeerreviewmaketitle

\vspace{-0.2cm}
\section{Introduction} \label{sec:introduction}
\jnlrev{Mobile edge devices (e.g., smartphones and wearables) increasingly demand intelligent language services directly on-device~\cite{fang2026dynamic, yi2025edgemoe, lin2026redundancy,hu2025agentscomerge, zhang2026resource,lin2025hsplitlora,fang2026hfedmoe}, driving the development of large language models (LLMs) such as GPT~\cite{achiam2023gpt}, LLaMA~\cite{touvron2023llama}, and DeepSeek~\cite{liu2024deepseek} that can model complex linguistic structures and generalize across diverse edge applications~\cite{lin2025pushing}.}
\peerrev{The success of these models is largely attributed to the scaling of model parameters and training data, which enables LLMs to learn rich semantic representations during pre-training and adapt effectively to downstream tasks through fine-tuning.
However, continuously scaling dense transformer models \jnlrev{leads to rapidly growing computational costs}~\cite{zeng2025efficientmoe, du2024sida, wang2024scaling}, \jnlrev{inherently conflicting with the resource constraints of edge devices.}
\rev{To address this limitation}, recent LLMs~\cite{lin2026moe, muennighoff2024olmoe, dai2024deepseekmoe, fedus2022switch} increasingly adopt Mixture-of-Experts (MoE) architectures}, which expands model capacity through a large set of specialized experts while activating only a sparse subset~\cite{dai2024deepseekmoe, fedus2022switch, lu2024not, fang2026hfedmoe, mei2024fedmoe, guo2021pfl}.
This sparse activation preserves most of the representational capacity of the model, allowing MoE models to achieve performance comparable to its large dense counterparts (e.g., LLaMA-2 7B) with substantially reduced computational cost~\cite{dai2024deepseekmoe}, making it a promising foundation for \jnlrev{deploying LLM services on mobile edge devices.}

\begin{figure}[t!]
\centering
\includegraphics[width=9cm]{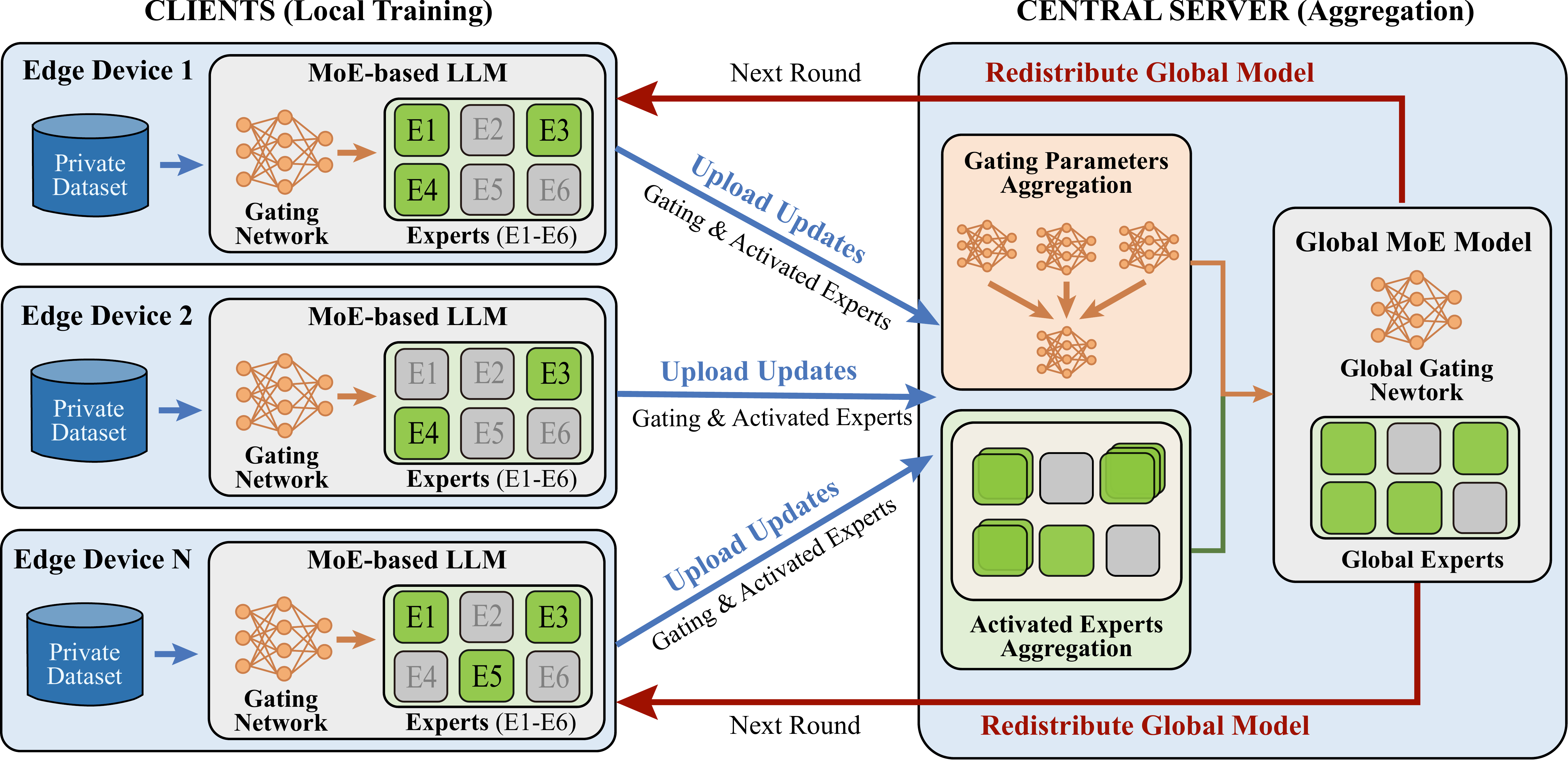}
\vspace{-3ex}
\caption{Federated MoE-based LLM fine-tuning workflow \jnlrev{in
edge computing systems}.}
\label{fig:fl_workflow}
\vspace{-2ex}
\end{figure}

\peerrev{In many real-world \jnlrev{mobile applications (e.g., personalized healthcare monitoring)}, the excessive data required for effective MoE-based LLM fine-tuning for downstream task adaptation raises serious privacy concerns, as sensitive information \jnlrev{(e.g., personal clinical data generated by smartphones)} cannot be shared across \jnlrev{edge networks} under strict privacy regulations, posing a substantial barrier to centralized fine-tuning.}
Federated learning (FL)~\cite{mcmahan17a, zongxuan2025communication, luo2023optimization, 
fang2025automated,hong2026conflict,teng2026communication,
xiongyue2024digital,lin2024fedsn} has therefore emerged as the paradigm for collaborative MoE fine-tuning across \jnlrev{edge devices} without exposing raw data,
reducing \jnlrev{on-device computational overhead} with sparse activations while preserving mobile user privacy.
In this workflow, as illustrated in Fig.~\ref{fig:fl_workflow}, each edge client locally activates and fine-tunes a sparse subset of experts selected by the gating network using its private datasets. The updated experts and gating parameters are then aggregated on the central server to construct an improved global MoE model, effectively integrating diverse knowledge \jnlrev{from distributed edge devices}.

While MoE-based LLM fine-tuning holds significant promise in  FL under \jnlrev{edge computing systems}, it encounters critical challenges due to inherent data heterogeneity across \jnlrev{edge devices}.
\jnlrev{In mobile computing systems, edge devices} always collect local data based on \jnlrev{personalized user contexts (e.g., exercise-related queries or chronic-disease-management queries)}, resulting in significant feature discrepancies and data not necessarily independent and identically distributed (non-IID)~\cite{zhu2021federated, wang2022clustered, xie2025dflmoe}.
\rev{Since MoE models rely on a gating network to dynamically route inputs to specialized experts~\cite{du2024sida, xie2025dflmoe}}, this data heterogeneity drives \jnlrev{edge devices} to learn highly divergent gating preferences that select client-specific expert subsets tailored to local data characteristics~\cite{lu2024not, mei2024fedmoe, guo2021pfl}, \jnlrev{hindering the efficiency of federated MoE-based LLM fine-tuning.} 
First, achieving global gating network that remains aligned with client-specific gating preferences is challenging.
Directly aggregating gating parameters \jnlrev{that optimized across disparate user contexts} can disrupt locally learned expert-activation patterns, making the global gating network poorly matched to clients' local gating preferences~\cite{xie2025dflmoe},
which leads to suboptimal expert selection \jnlrev{that forces the device to activate irrelevant experts, thereby wasting edge computing resources and degrading local performance.}
Second, data heterogeneity poses significant challenges in preserving expert specialization in \jnlrev{mobile federated MoE fine-tuning}. 
As local experts are updated on non-IID data \jnlrev{from distinct mobile clients}, the same-indexed expert is optimized under different conditional feature–label correlations, causing it to learn distinct semantics across clients. 
Aggregating these semantically inconsistent experts blurs its semantics and undermines specialization, resulting in unstable convergence and degraded performance \jnlrev{that severely impairs convergence efficiency and exacerbates mobile network overhead.}
We will empirically conduct measurement studies in Sec.~\ref{sec:motivation} to investigate these challenges.


To tackle the above challenges, we propose a \underline{fed}erated aggregation \underline{align}ment framework for \underline{MoE}-based LLM fine-tuning, named \name, \waitrev{which 
enforces routing consistency across locally optimized gating behaviors and semantic alignment among same-indexed experts under heterogeneous client data.}
First, to resolve gating preference misalignment across clients, \peerrev{instead of enforcing parameter-level consensus \jnlrev{that degrades local on-device performance}, we propose a consistency-based gating distribution alignment mechanism that align clients by constructing a global routing distribution and then regularize the local gating networks towards this referenced distribution, preserving each \jnlrev{edge device}’s discriminative local preference for expert selection while promoting consistent cross-client gating behaviors.}
Second, \rev{since aligning routing behaviors alone cannot guarantee that experts with the same index maintain consistent semantic roles across clients}, we further introduce a semantic-aware expert aggregation strategy that explicitly quantifies the semantic consistency of same-indexed experts across clients. Based on a region-conditioned gate aggregation with an adaptive threshold, \name selectively aggregates updates from semantically aligned clients while suppressing conflicting ones, \peerrev{thereby guiding experts with the same index to converge toward consistent semantic roles while preserving their specialization across clients.}
The key contributions of this paper are summarized as follows.
\begin{itemize}
  \item \waitrev{We propose \name, a federated MoE-based LLM aggregation framework \jnlrev{tailored for mobile edge environments} to align expert and gating network behaviors across clients, enhancing LLM local fine-tuning performance under heterogeneous data distributions.}
  \item We design a consistency-based gating distribution alignment scheme that aligns routing output distributions based on consistency weighting and constrains local gating networks during fine-tuning, maintaining cross-client consistency without overriding client-specific preference.
  \item We develop a semantic-aware expert aggregation strategy to adaptively align updates of same-indexed experts from semantically consistent clients, thereby stabilizing expert specialization while promoting global convergence.
  \item We conduct extensive experiments to evaluate the performance of \name, which demonstrates its superiority in \jnlrev{model accuracy and convergence acceleration with communication and computational efficiency}, proving its deployability for mobile edge computing scenarios.
\end{itemize}

This paper is organized as follows.
Sec.~\ref{sec:motivation} motivates the design of \name by revealing the challenges of incorporating MoE into FL in edge computing systems, and Sec.~\ref{sec:design} elaborates on the framework design. Simulation setup and performance evaluation are presented in Sec.~\ref{sec:simulation_simu_setup} and Sec.~\ref{sec:simulation}.
Related works and technical limitations are discussed in Sec.~\ref{sec:related_work}, followed by conclusions in Sec.~\ref{sec:conclusion}.

\section{Challenges and Motivation} \label{sec:motivation}
In this section, we conduct extensive pilot studies to elaborate the key challenges of fine-tuning MoE-based LLM in \jnlrev{mobile federated learning systems}, which motivates the design of our \name.

\begin{figure*}[t]
\vspace{-1ex}
\centering
\subfloat[Gating distribution in Client 1]{
\includegraphics[width=0.235\linewidth]{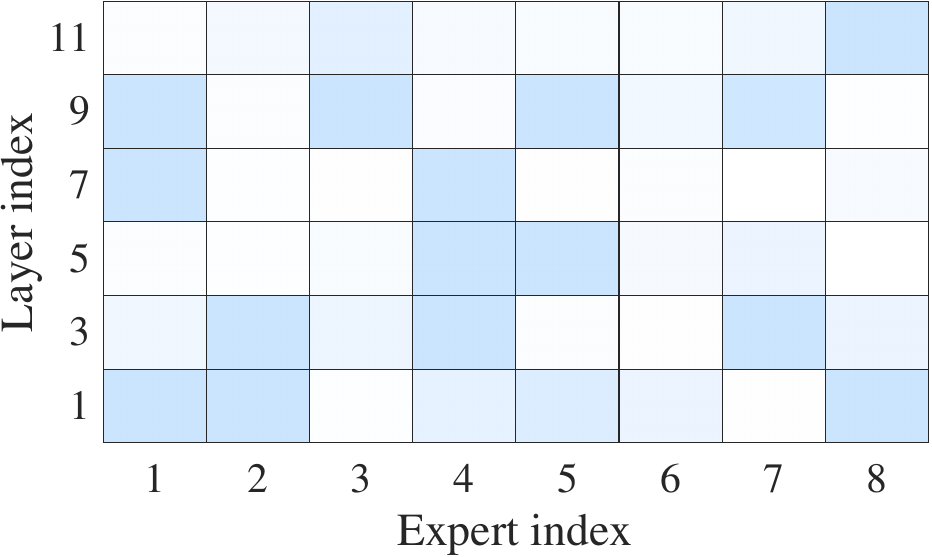}
}
\subfloat[Gating distribution in Client 2]{
\includegraphics[width=0.235\linewidth]{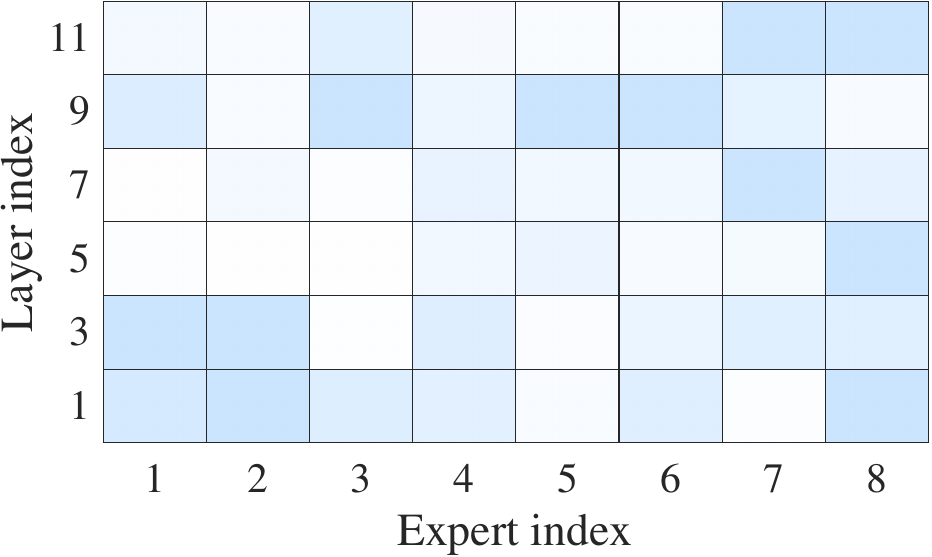}
}
\subfloat[Gating distribution in Client 3]{
\includegraphics[width=0.235\linewidth]{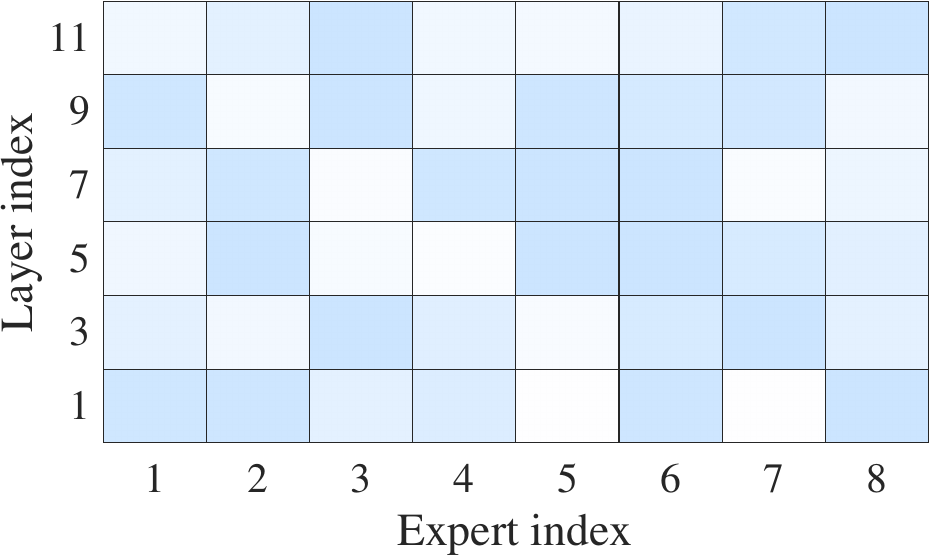}
}
\subfloat[Expert selection across clients \label{fig:mtv_heterogeneity_selection}]{
\includegraphics[width=0.235\linewidth]{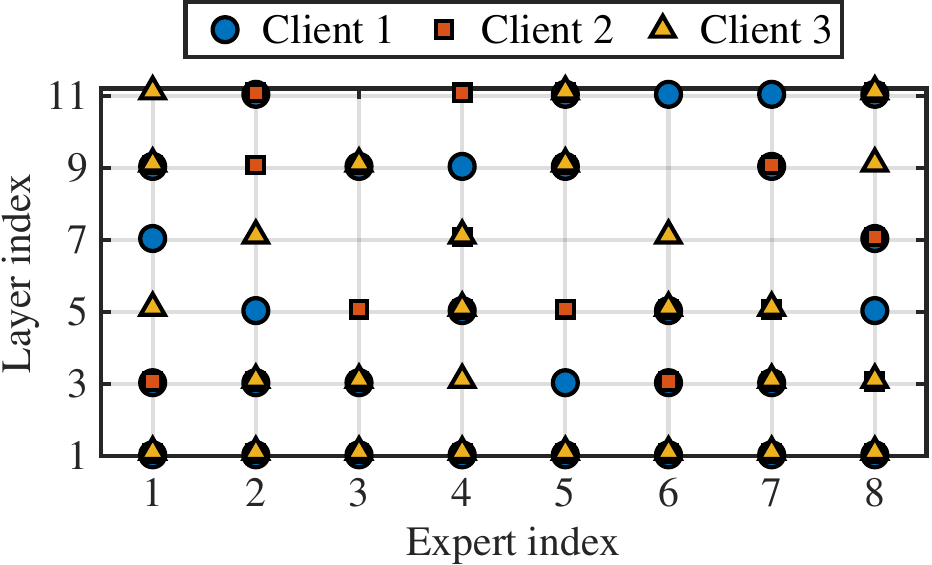}
}
\vspace{-1ex}
\caption{The expert selection preferences across \jnlrev{edge devices} before aggregation under non-IID data distributions.}
\label{fig:mtv_heterogeneity}
\vspace{-6ex}
\end{figure*}

\begin{figure}[t]
\centering
\subfloat[Visualization of gating parameters \label{fig:mtv_gating_overlap}]{
\includegraphics[width=0.49\columnwidth, height=2.6cm]{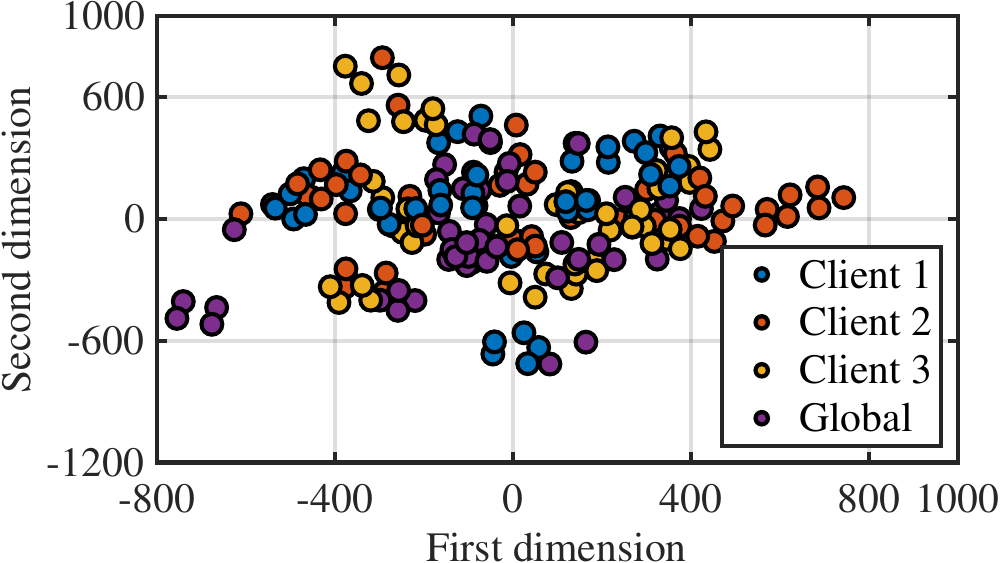}
}
\subfloat[Expert selection after aggregation \label{fig:mtv_gating_performance}]{
\includegraphics[width=0.51\columnwidth, height=2.6cm]{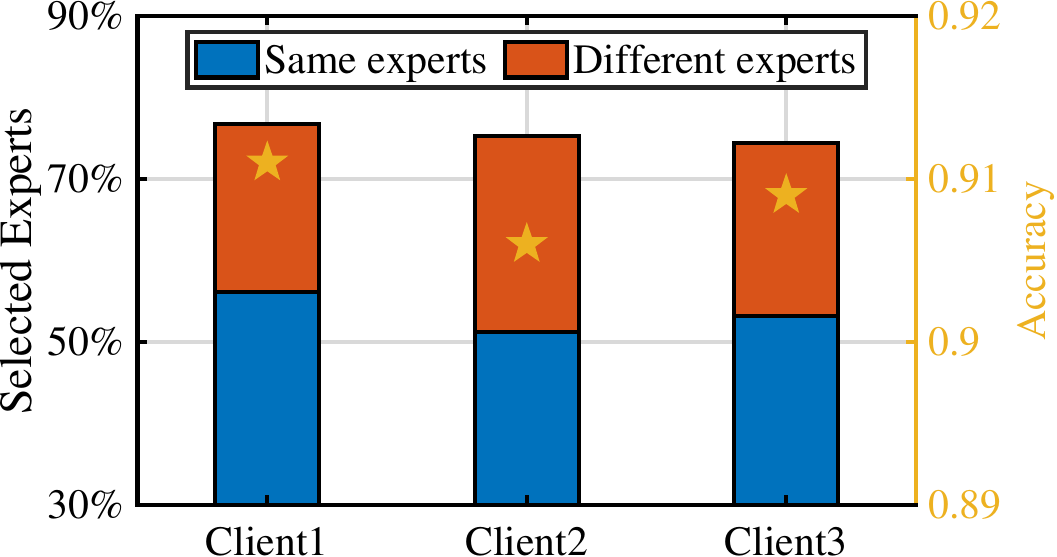}
}
\vspace{-0.5ex}
\caption{The t-SNE visualization of gating parameters and the performance of expert selection from direct parameter aggregation across \jnlrev{mobile clients} on AGNews dataset.}
\label{fig:mtv_gating}
\vspace{-2ex}
\end{figure}

\vspace{-1.5ex}
\subsection{Gating Preference Misalignment} \label{sec:mtv_gating}
In \jnlrev{mobile} federated MoE systems, each \jnlrev{edge device} independently optimizes its gating network to determine which experts should be activated based on heterogeneous local data, leading to client-specific expert routing decision boundaries~\cite{yi2026pfedmoe, feng2025pm}.
However, during global aggregation, existing methods~\cite{guo2021pfl, zec2020specialized, feng2025pm, zhan2024fedmoe, yi2026pfedmoe, mei2024fedmoe, fang2026hfedmoe} usually combine these gating networks through parameter averaging across clients, \waitrev{which linearly blends incompatible routing decisions into a unified but distorted model, thereby entangling diverse expert selection preferences learned across \jnlrev{different mobile users.}}
This mixture makes the global gating network \waitrev{tend to misalign with any \jnlrev{edge device}’s data characteristics~\cite{xie2025dflmoe}.}
As a result, inputs that previously activated the most suitable experts on each client may be redirected to less relevant ones after aggregation, thereby degrading local performance and \jnlrev{wasting the limited computing budgets of edge devices on sub-optimal experts.}

To investigate the impact of aggregating gating parameters on expert selection preferences, we conduct motivating studies using Switch Transformer~\cite{fedus2022switch} with 8 experts per layer on the AGNews dataset~\cite{zhang2015character}, where each \jnlrev{edge device} independently fine-tunes gating network on local non-IID data and evaluates the performance on the same test dataset.
Fig.~\ref{fig:mtv_heterogeneity} demonstrates that locally optimized gating networks exhibit distinct preferences, leading to client-specific expert selections (see Fig.~\ref{fig:mtv_heterogeneity_selection}) that closely reflect local data characteristics. 
In contrast, Fig.\ref{fig:mtv_gating_overlap} reveals that directly averaging gating parameters across \jnlrev{edge devices} \waitrev{disrupts learned routing decisions, yielding a global gating network whose expert selection is misaligned with all clients}, thereby resulting in notable performance degradation after aggregation as shown in Fig.~\ref{fig:mtv_gating_performance}.
These observations underscore the necessity of explicitly aligning \needrev{gating output distributions}, rather than parameters, to preserve client-specific expert selection preferences and thereby improve local performance \jnlrev{under mobile computing constraints}.

\begin{figure}[t]
\centering
\subfloat[Representations of experts in the \\last layer on client 1 \label{fig:mtv_expert_semantic}]{
\includegraphics[width=0.49\linewidth]{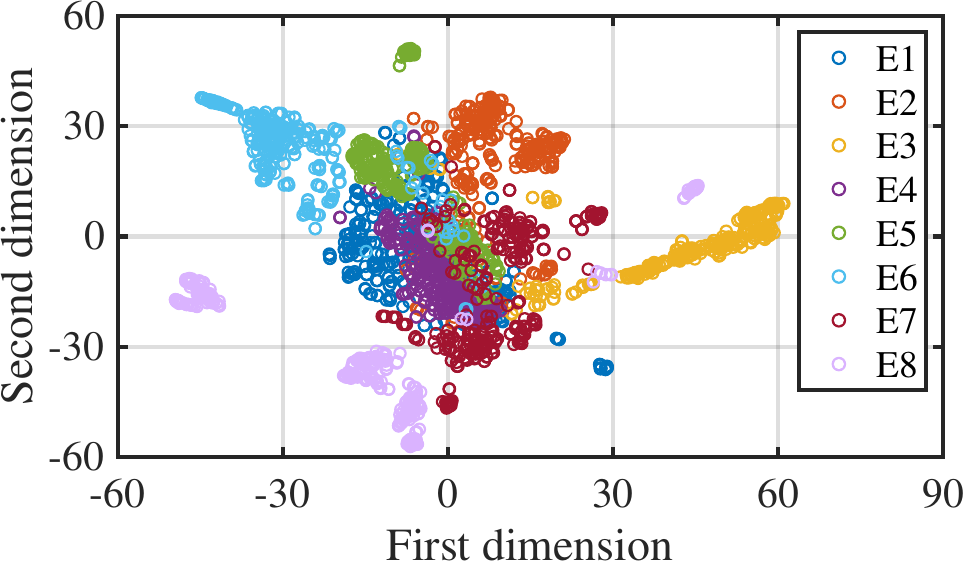}
}
\subfloat[Representations of experts in the \\last layer on client 2 \label{fig:mtv_expert_semantic}]{
\includegraphics[width=0.49\linewidth]{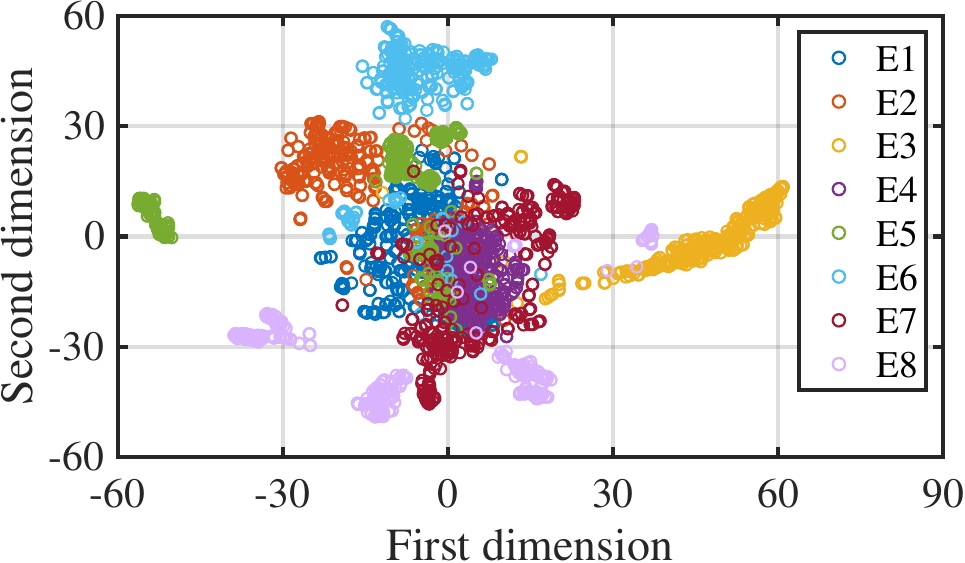}
}\\
\vspace{-1ex}
\subfloat[Output representation distributions of \\experts in the last layer on client 3 \label{fig:mtv_expert_semantic}]{
\includegraphics[width=0.49\linewidth]{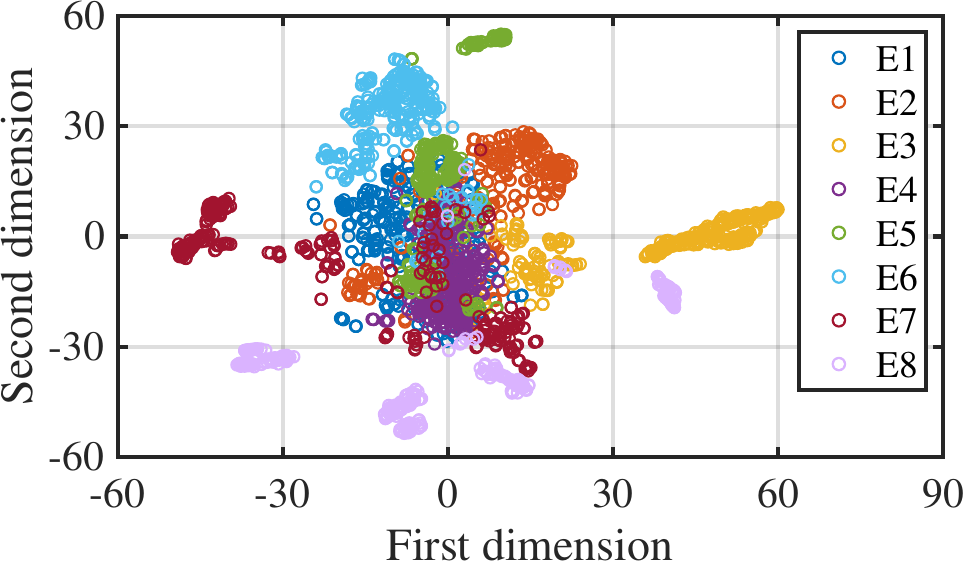}
}
\subfloat[Local performance on client-specific data before or after aggregation \label{fig:mtv_expert_performance}]{
\includegraphics[width=0.49\linewidth]{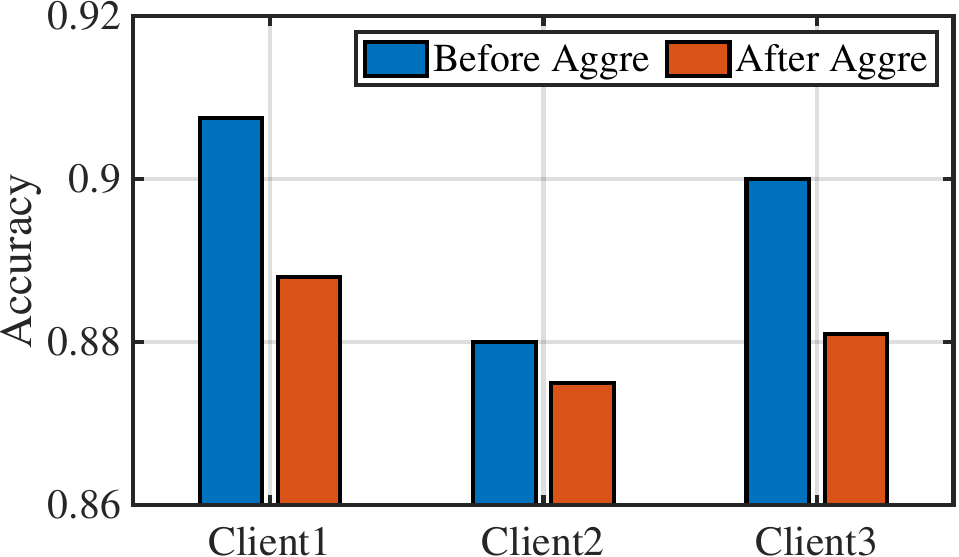}
}
\vspace{-0.5ex}
\caption{The t-SNE representation visualization and local performance for expert specialization on AGNews dataset.}
\label{fig:mtv_expert}
\vspace{-2ex}
\end{figure}

\vspace{-1ex}
\subsection{Expert Semantic Divergence} \label{sec:mtv_expert}
MoE-based LLM fine-tuning typically uses gating networks to partition input space into several regions, with each expert responsible for a specific region~\cite{hu2025fft}.
Although effectively mitigating data heterogeneity in centralized MoE through a globally consistent input-space partition, it fails to maintain expert specialization across \jnlrev{distributed mobile clients}. 
In federated MoE fine-tuning, each expert is optimized under non-IID conditional data distributions with client-specific characteristics~\cite{lu2024not, guo2021pfl}, \waitrev{where the discrepancy in feature–label correlations within the same input region drives a single expert toward distinct input–output mappings on different \jnlrev{edge devices}, causing the same indexed expert to represent different semantic regions.} Directly aggregating such semantically misaligned expert parameters forces a single expert to serve the union of incompatible regions, which destroys expert specialization and blurs expert semantics, thus resulting in unstable convergence \jnlrev{that  exacerbates mobile network communication overhead.}


To empirically demonstrate the expert semantic divergence, \needrev{we compare the output distribution and performance of the same indexed expert across clients in locally fine-tuned MoE} on the AGNews dataset using Switch Transformer with 8 experts. We train a global MoE and then freeze its gating parameters on all clients to eliminate routing discrepancy, while allowing each expert to be updated locally using client-specific non-IID data distributions.
It is illustrated in Fig.~\ref{fig:mtv_expert} that the same indexed expert exhibits large inter-client divergences in \rev{output embedding distributions} on the same test dataset, indicating that it converges to distinct semantic region after local fine-tuning on different \jnlrev{edge devices}. 
Fig.~\ref{fig:mtv_expert_performance} further shows that directly aggregating these semantically misaligned expert parameters leads to inferior local performance, 
necessitating semantic-aware expert aggregation to detect inter-client semantic mismatch and selectively combine expert updates for specialization preservation.


\begin{figure*}[t]
\vspace{-1ex}
\centering
\includegraphics[width=14.5cm]{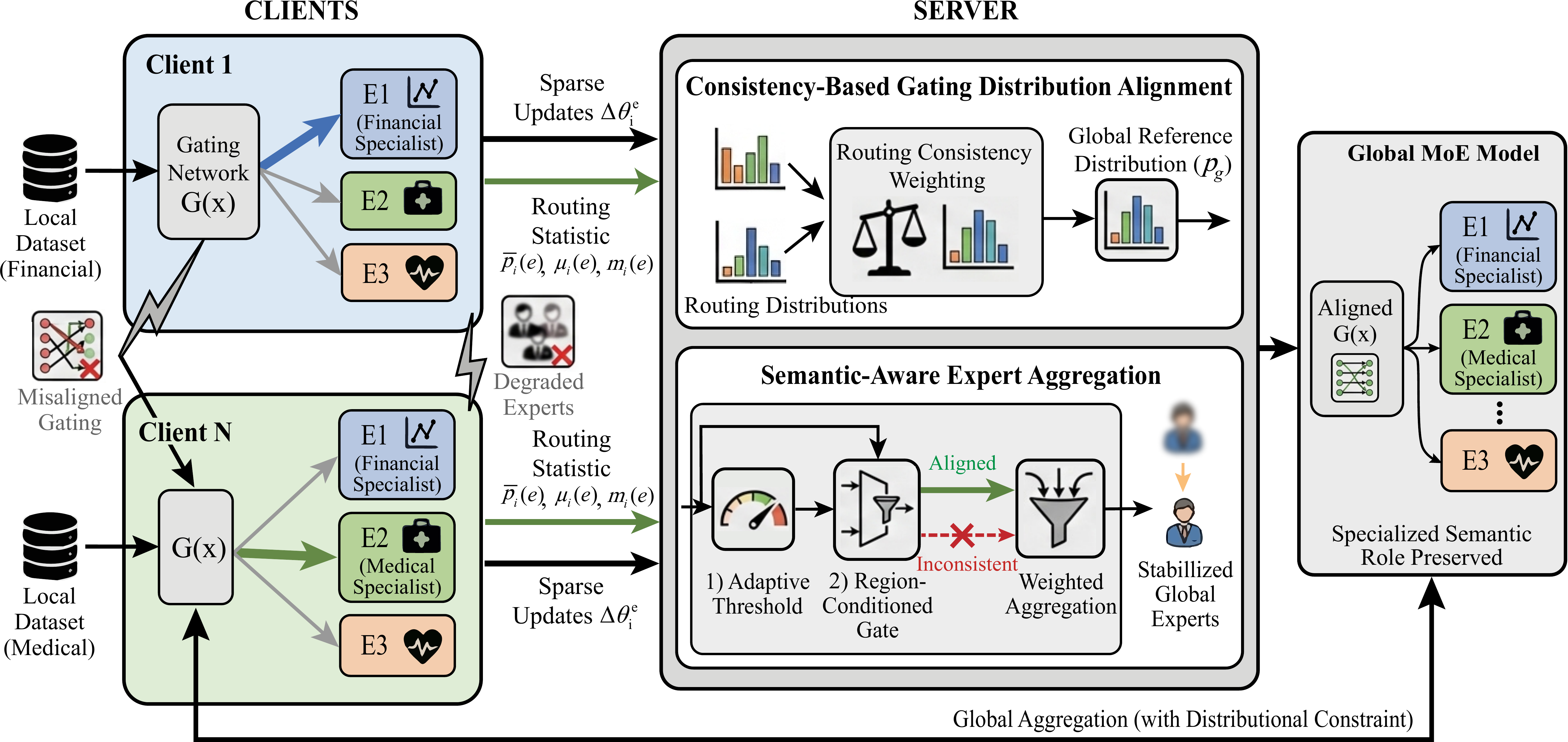}
\caption{An overview of \name framework, where the server aligns routing behaviors across clients via consistency-aware alignment of routing distributions and performs semantic-aware expert aggregation to stabilize the semantics of the same-indexed experts in the global MoE model.}
\label{fig:framwork}
\vspace{-2ex}
\end{figure*}

\section{Framework Design}\label{sec:design}
In this section, we introduce the design of \name.
We will outline the system overview and workflow, followed by a detailed description of the \name framework.

\vspace{-1ex}
\subsection{Overview}\label{sec:dsn_overview}
As shown in Fig.~\ref{fig:framwork}, \name consists of two core components: consistency-based gating distribution alignment and semantic-aware expert aggregation.
First, to resolve gating preference misalignment across clients, we introduce a consistency-based gating distribution alignment mechanism that
first constructs a global routing reference by aggregating gating distributions (Sec.~\ref{sec:dsn_gating_aggregation}) with a routing consistency weighting scheme to measure client contributions (Sec.~\ref{sec:dsn_gating_weighting}), and then optimizes local gating behaviors through distribution regularization with this global reference during fine-tuning (Sec.~\ref{sec:dsn_gating_regularization}), enabling cross-client routing consistency with preserved discriminative local preferences.
Second, to mitigate expert semantic divergence, the semantic-aware expert aggregation module selectively aggregates expert updates from semantically consistent clients while suppressing conflicting ones (Sec.~\ref{sec:dsn_expert_aggregation}), thereby stabilizing the semantic roles of experts across clients and accelerating global MoE convergence. To further enhance expert specialization, an adaptive threshold is introduced into the aggregation weights to dynamically calibrate alignment sensitivity for different experts (Sec.~\ref{sec:dsn_expert_threshold}).

The workflow of \name follows three steps: 
i) During local fine-tuning on private, non-IID datasets, each client updates its activated experts and aligns local gating behaviors with a global reference distribution.
ii) In the end of a training round, each client computes the required metrics for expert-level aggregation. \jnlrev{Instead of uploading the entire local model with massive communication overhead}, only the routing output distributions, the activated experts, and their associated statistics are uploaded to the server for aggregation. 
iii) The server then performs consistency-based gating distribution alignment to construct an updated global reference, and applies the semantic-aware expert aggregation for each expert index based on the reported statistics. Finally, the global gating distribution and aggregated experts are redistributed to all clients for the next training round.

\vspace{-1ex}
\subsection{System Model} \label{sec:dsn_model}
\vspace{-0.5ex}
We consider a \jnlrev{mobile edge} federated learning system with a central server and $N$ \jnlrev{mobile clients} collaboratively fine-tuning a global MoE-based LLM under heterogeneous local data distributions. Each client $i \in \{1, \dots, N\}$ holds a private local dataset $\mathcal{D}_i$ drawn from a distinct distribution $\mathcal{P}_i$. The data is non-IID due to the heterogeneous nature of \jnlrev{local device environments}, that is, $\mathcal{P}_i$ and $\mathcal{P}_j$ may not be the same for different client $i$ and $j$. 
The collaborative objective is to minimize the global loss function $\mathcal{L}(\Theta) = \sum_{i=1}^N \frac{|\mathcal{D}_i|}{|\mathcal{D}|} \mathcal{L}_i(\Theta_i)$, where $|\mathcal{D}|$ is the total number of samples, and $\mathcal{L}_i(\Theta_i) = \mathbb{E}_{x \sim \mathcal{P}_i} [\ell(x; \Theta_i)]$ represents the local objective with $\ell(\cdot)$ being the task-specific loss function and $\Theta_i$ the parameters of local model.
In the MoE-based LLM architecture, the backbone model $\Theta_i$ at each client $i$ contains $L$ MoE layers that replace the standard dense feed-forward networks. Each MoE layer consists of a gating network and $S$ parallel experts, where the expert set is denoted by $\mathcal{E} = \{1, ..., e, ..., S\}$. 
\peerrev{Given an input $x$, the gating network in each layer computes routing scores $G_i(x)$ over experts, and only the top-$k$ experts ($k \ll S$) with the highest routing scores are selected for fine-tuning, forming the activated expert set $\mathcal{E}_i = \text{Top-}k\left(\{G_i(x)\}_{e \in \mathcal{E}}\right)$.
The routing scores of the selected experts are then normalized to the routing probability vector $\mathbf{p}_i(x)=[\mathbf{p}(x,1), ..., \mathbf{p}(x,e), ..., \mathbf{p}(x,S)]$ as}
\begin{equation}
\mathbf{p}_i(x) = \text{Softmax}\left(G_i(x)_{e \in \mathcal{E}_i}\right).
\end{equation}
The final output of the MoE layer is computed as the weighted sum of the activated experts' outputs:
\begin{equation}
y_i = \sum_{e \in \mathcal{E}_i} \mathbf{p}_i(x,e) \cdot \mathbf{E}_e(x; \theta_{i}^e),
\end{equation}
where $\mathbf{E}_i(\cdot)$ and $\theta_i^e$ denote the output function and the parameters of expert $e$, respectively. 
During local training round $t$, client $c$ fine-tunes the local MoE-based LLM with sparse activation, updating only the gating parameters $\theta_i^g$ and the subset of experts $\mathcal{E}_i \subseteq \mathcal{E}$ activated by its local data $\mathcal{D}_i$. After local training, the client uploads the gating output distributions and activated expert weights $\{\theta_i^e\}_{e \in \mathcal{E}_i}$ to the server for aggregation.

\vspace{-0.5ex}
\subsection{Consistency-based Gating Distribution Alignment}
\label{sec:dsn_gating}
\begin{figure}[t!]
\centering
\includegraphics[width=8.5cm]{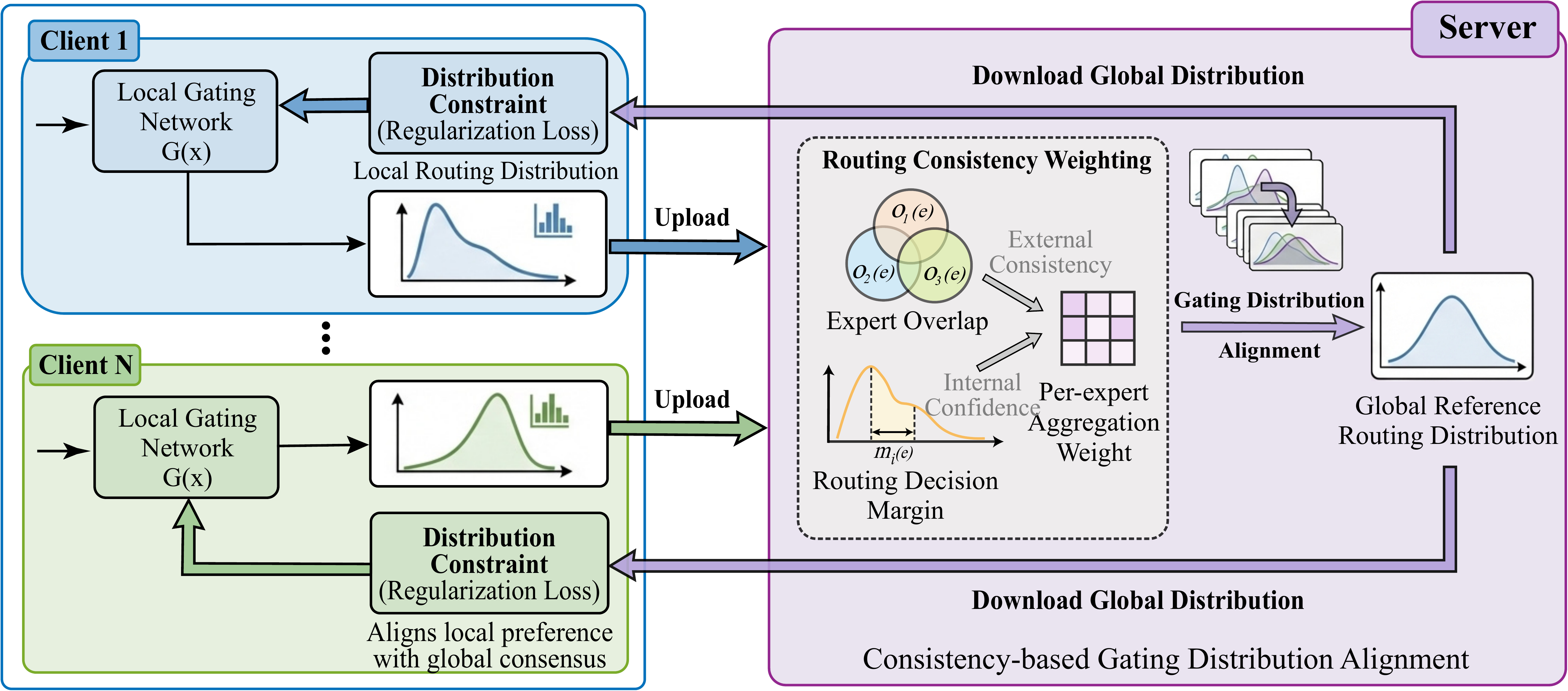}
\caption{The gating distribution alignment with routing consistency weighting during global aggregation and \rev{adaptive} distribution regularization during local fine-tuning.}
\label{fig:gating_alignment}
\vspace{-2ex}
\end{figure}

As discussed in Sec.~\ref{sec:mtv_gating}, gating parameter aggregation in existing MoE-based FL studies~\cite{guo2021pfl, zec2020specialized, feng2025pm, zhan2024fedmoe, yi2026pfedmoe, mei2024fedmoe, fang2026hfedmoe} without accounting for client-specific expert selection preferences may fail to keep alignment with clients' data characteristics, which severely degrades local performance under non-IID data distributions \jnlrev{and causes edge devices to waste computing resources activating suboptimal experts.}
To mitigate this gating misalignment, we move from parameter-space aggregation to \peerrev{function-space alignment on gating behaviors. 
\needrev{Since output distributions of the local gating network represents selection probabilities of each expert that represents gating preferences,} aligning these output distributions encourages clients to agree on consistent expert selection preferences while allowing distributional variations that reflect discriminative client-specific preference. 
Therefore, we introduce a consistency-based gating distribution alignment strategy, as illustrated in Fig.~\ref{fig:gating_alignment}, which aligns behavioral consensus of gating networks through expert-level routing consistency weighting $\omega_i(e)$ during aggregation, rather than forcing uniform gating parameters $\theta_i^g$ across clients.} 
\waitrev{During local fine-tuning, each client aligns its gating behavior $\mathbf{p}_i(x,e)$ with the global gating distribution $\mathbf{p}_g(e)$ to enhance consistency in shared data regions, while retaining flexibility in client-specific regions with low agreement, allowing expert selection to remain sensitive to client-specific data distributions.}

\subsubsection{\textbf{Gating Distribution Alignment}}
\label{sec:dsn_gating_aggregation}

Since all clients share an identical gating architecture and select experts from the same expert set, their routing output distributions reside in a common probability space and provide a direct and comparable representation of expert selection behavior across clients.
These distributions encode client-specific expert selection preferences shaped by local data, capturing which experts are suitable for each input and the confidence with which the gating network discriminates among experts.
Therefore, under data heterogeneity, discrepancies in routing output distributions explicitly reflect divergences in such preferences across clients.
\waitrev{Aligning routing output distributions, rather than aggregating parameters, operates on the functional mapping from inputs to expert selection behaviors, ensuring that alignment is performed on routing decisions learned from each client’s data instead of on unconstrained parameter representations,
thereby making it a promising target for cross-client gating alignment.}

Based on this insight, we align clients' expert selection preferences by enforcing consistency in their routing output distributions rather than directly matching gating parameters. 
%
Specifically, during each local training round $t$, client $i$ computes the empirical average routing distribution $\bar{\mathbf{p}}_i(e)$ of expert $e$ over its local dataset $\mathcal{D}_i$ as
\begin{equation}
\bar{\mathbf{p}}_i(e) = \frac{1}{|\mathcal{D}_i|} \sum_{x \in \mathcal{D}_i} \mathbf{p}_i(x,e).
\end{equation}

Instead of uploading gating network parameters, each client only uploads these per-expert activation distributions to the server \jnlrev{with minimal communication overhead}.
\peerrev{The server then aggregates these routing statistics 
and the global routing distribution for each expert is
\begin{math}
\mathbf{p}_g(e) = \frac{1}{N} \sum_{i=1}^N \bar{\mathbf{p}}_i(e),
\end{math}
which \needrev{reflects global consensus on the routing decision for expert $e$}, serving as a global reference to guide client-side gating behavior optimization.}


\subsubsection{\textbf{Routing Consistency Weighting}}
\label{sec:dsn_gating_weighting}
To derive a global routing distribution $\mathbf{p}_g(e)$ that remains consistent across client-specific preferences while preserving their locally learned \needrev{expert-activation patterns}, \peerrev{the key challenge lies in how to appropriately weight client contributions as uniform aggregation may dilute consistent cross-client selections while override locally discriminative preferences.}
We therefore design a routing consistency weighting scheme for aggregating client-specific routing distributions, which assigns higher aggregation weights to clients whose gating behaviors are both externally consistent and internally confident, such that global routing decisions reflect stable expert activations broadly applicable across clients, without suppressing expert selections relevant to local data.
Specifically, we quantify each client's routing consistency using two metrics: i) expert overlap, which measures the external consistency between local routing distribution and global gating behavior to promote stable expert activations; and ii) routing decision margin, which reflects the internal confidence of a client's routing decisions to better preserve local selection preferences.

\textbf{Expert overlap.}
To establish a representative global routing distribution that captures generalizable behavior, it is essential to prioritize experts exhibiting broad activation across clients.
To this end, we introduce an expert overlap metric to measure the consistency between client $i$'s specific selection preference and the cross-client behavior, providing a reliable indicator of global consensus.
By prioritizing globally important experts and shared routing decisions, clients whose routing distributions substantially deviate from cross-client behavior receive reduced weights. This prevents client-specific or incompatible boundaries \jnlrev{caused by highly anomalous mobile environments} from dominating global aggregation, thus maintaining consistent routing decisions across clients. 


\peerrev{Since effective overlap exists only when local gating preference and global consensus mutually validate each other}, we formulate the expert overlap metric $o_{i}(e)$ for expert $e$ as the product of local and cross-client routing distributions:
\begin{equation}
o_{i}(e) = \bar{\mathbf{p}}_{i}(e) \cdot \frac{1}{N-1}\sum_{j=1, j\ne i}^N \bar{\mathbf{p}}_{j}(e),
\end{equation}
where $\bar{\mathbf{p}}_{i}(e)$ is the empirical average routing distribution of client $i$ and $ \frac{1}{N-1}\sum_{j\ne i} \bar{\mathbf{p}}_{j}(e)$ denotes the global routing consensus. 
\waitrev{The expert overlap captures the external consistency of a client's gating behavior: when a client frequently activates $e$ ($\bar{\mathbf{p}}_{i}(e)$ is large) while the activation probability on other clients remains low ($\frac{1}{N-1}\sum_{j\ne i} \bar{\mathbf{p}}_{j}(e)$ is small), the resulting small overlap identifies this usage pattern as \jnlrev{highly
specific to a single mobile user} rather than globally shared, \rev{thereby preventing such localized behaviors from exerting disproportionate influence during aggregation.}}

\textbf{Routing decision margin.}
While the expert overlap metric $o_i(e)$ identifies experts that are widely activated across clients, the client-specific patterns captured by certain experts may disproportionately influence the resulting global distribution.
Therefore, we utilize the routing decision margin to quantify the internal confidence of each client, preserving reliable local decision boundaries while preventing corruption of the global routing distribution by inconsistent clients.
Specifically, clients exhibiting large decision margins (e.g., consistently activate the same expert for similar inputs) contribute more to the global routing distribution, as such behavior reflects a stable and well-formed decision boundary.
In contrast, the influence of clients with narrow margins or inconsistent top-$k$ expert selections is down-weighted to shield the global distribution from unreliable or noisy routing decisions.

For each expert $e$, we define the routing decision margin over client $i$'s local dataset $\mathcal{D}_i$ as follows:
\begin{equation}
m_{i}(e)=\frac{1}{|\mathcal{D}_i|}\sum_{x\in\mathcal{D}_i} 
\max\{0,\, \mathbf{p}_i(x,e)-\max_{e'\neq e} \mathbf{p}_i(x,e')\},
\end{equation}
where $\mathbf{p}_i(x,e)-\max_{e'\neq e} \mathbf{p}_i(x,e')$ reflects the decision dominance with which it surpasses alternative experts.
Large margins indicate that client $i$ maintains a stable and confident decision boundary for expert $e$, whereas small margins reflect ambiguous or unstable gating behavior. Zero margins correspond to experts that are never selected and should therefore exert negligible influence during aggregation.

Building on the complementary roles of external consistency $o_i(e)$ and internal confidence $m_i(e)$, we compute a per-expert aggregation weight as
\begin{equation} \label{eq:routing_consist_weight}
\omega_i(e) = \frac{s_i(e)}{\sum_{j=1}^{N} s_j(e)}, s_i(e) = o_i(e) \cdot m_i(e),
\end{equation}
where the multiplicative design enforces a joint-reliability constraint that prioritizes clients whose routing decisions for expert $e$ are both internally stable and externally aligned \rev{with the federation}, thereby preserving both global consistency and client-specific decision boundaries.

\peerrev{With the aggregation weights $\omega_i(e)$ for each client $i$, the global routing consistent distribution $\mathbf{p}_g(e)$ for each expert can be rewritten as
\vspace{-1ex}
\begin{equation}
\mathbf{p}_g(e) = \sum_{i=1}^N \omega_i(e) \bar{\mathbf{p}}_i(e),
\end{equation}
which serves as the global reference for client-side routing distribution alignment, bridging individual decision boundaries with collective knowledge.}



\subsubsection{\textbf{Client-Side Optimization with Distribution Regularization}}
\label{sec:dsn_gating_regularization}
To bridge the gap between divergent local and global routing stability, probabilistic alignment optimizes expert selection directly in the functional space without imposing rigid parameter constraint.
Upon obtaining the aggregated global consistent routing distribution $p_g(e)$, we introduce a client-side optimization strategy that aligns local gating behaviors with this global reference through distribution regularization.
Distinct from existing frameworks~\cite{guo2021pfl, zec2020specialized, feng2025pm, zhan2024fedmoe, yi2026pfedmoe, mei2024fedmoe, fang2026hfedmoe} that rely on gating parameter aggregation, this probabilistic alignment optimizes expert selection directly in the functional space.
\peerrev{By employing Kullback–Leibler (KL) divergence as a soft-constraint regularization for local routing distribution, the tolerance for deviation on client-specific preferences is achieved through probabilistic similarity of the objective, which allows each local model to preserve its own parameter structure while steering its gating behavior toward the global consensus.}
Therefore, for each input $x$, each client aligns its local routing distribution $\mathbf{p}_i(x,e)$ with global routing target $\mathbf{p}_g(e)$ by minimizing a regularization loss:
\begin{equation}
\mathcal{L}_{\text{reg}} = \sum_{e \in \mathcal{E}} \alpha_i(e) \cdot \mathbb{D}_{KL}(\mathbf{p}_i(x,e) \| \mathbf{p}_g(e)),
\end{equation}
where $D_{\text{KL}}(\cdot)$ denotes the Kullback–Leibler divergence to measure distribution similarity and $\alpha_i(e)$ represents the adaptive weight for each expert, guiding local gating network toward \rev{global routing consensus} while preserving flexible adaptation to local expert selection preferences.

To preserve flexibility in weak-consensus regions while enforcing consistency on \needrev{shared experts}, we propose an adaptive weight strategy that dynamically adjusts the regularization strength based on \rev{expert-level consensus} rather than applying uniform alignment constraint across all experts.
Experts with high cross-client consistency are strongly aligned to improve cross-client generalization, whereas experts that only activated frequently on specific clients are weakly regularized to preserve client-specific selection preferences driven by local data characteristics.
\waitrev{Since the proposed expert overlap metric $o_i(e)$ serves as a proxy to quantify this consensus}, we define the adaptive weight for expert $e$ on client $i$ as:
\begin{equation} \label{eq:regular_alpha}
\alpha_i(e) = \sigma(o_i(e) - \eta),
\end{equation}
where $\eta$ represents the overlap threshold reflecting the global consistency of expert $e$ and $\sigma(\cdot)$ denotes the sigmoid function to provide a smooth and bounded transition from weak to strong alignment. 
Globally consistent experts ($o_i(e) \ge \eta$) receive large weights and thus strong alignment, while locally used experts are weakly regularized, allowing deviation from global gating behavior to preserve specialized preferences.

The global routing distribution is typically a smoothed aggregation of distinct client-specific preferences, whereas only Top-$k$ experts are activated for each input. \waitrev{To produce discriminative gating network while preserving the sparsity of MoE model}, we restrict the distribution regularization to a masked subset of experts that are either selected for current input or globally prominent.
Thus, the regularization loss $\mathcal{L}_{\text{reg}}$ can be rewritten as
\begin{equation}
\begin{aligned}
\mathcal{L}_{\text{reg}} &= \sum_{e \in \mathcal{E}} \alpha_i(e) \cdot \mathbb{D}_{KL}(\mathbf{p}_i(x,e) \| \mathbf{p}_g(e)) \\
&\cdot \mathbf{1}[e \in \text{Top-}k(\mathbf{p}_i(x,e)) \cup \text{Top-}k(\mathbf{p}_g(e))].
\end{aligned}
\end{equation}

Finally, each client jointly optimizes the local MoE training loss $\mathcal{L}_{\text{local}}$ (e.g., task-specific cross-entropy) and the regularization loss $\mathcal{L}_{\text{reg}}$. The local objective can be expressed as
\begin{equation} \label{eq:local_objective}
\mathcal{L}_{\text{total}} = \mathcal{L}_{\text{local}} + \lambda \mathcal{L}_{\text{reg}},
\end{equation}
where $\lambda$ is a balancing coefficient controlling the trade-off between task adaptation and gating consistency. 
\subsection{Semantic-Aware Expert Aggregation} \label{sec:dsn_expert}
Recalling Sec.~\ref{sec:mtv_expert}, data heterogeneity often drives identically indexed experts to represent semantically distinct input regions and divergent feature–label correlations across clients. Aggregating these experts forces them to handle a union of incompatible regions \jnlrev{across the edge network}, thereby compromising expert specialization and model sparsity.
While the gating distribution alignment proposed in Sec. \ref{sec:dsn_gating} enforces consistent cross-client expert selection preferences for similar inputs, 
\waitrev{it only regulates which expert is activated, rather than how that expert transforms the input. Since expert updates to feature–label mappings remain governed by client-specific data distributions, the semantic consistency on same indexed experts across clients is still not guaranteed.}
\peerrev{To bridge this gap, we propose a semantic-aware expert aggregation mechanism that quantifies the semantic similarity ($\mu_i(e)$, $D_{i,j}(e)$) across clients and selectively aligns expert updates from semantically consistent clients with $\gamma_{i,j}(e)$}, stabilizing the semantic roles of experts in \jnlrev{mobile clients} and accelerating global MoE convergence under heterogeneous data distributions.


\subsubsection{\textbf{Region-Conditioned Semantic Aggregation}}
\label{sec:dsn_expert_aggregation}
\begin{figure}[t!]
\centering
\includegraphics[width=8.5cm]{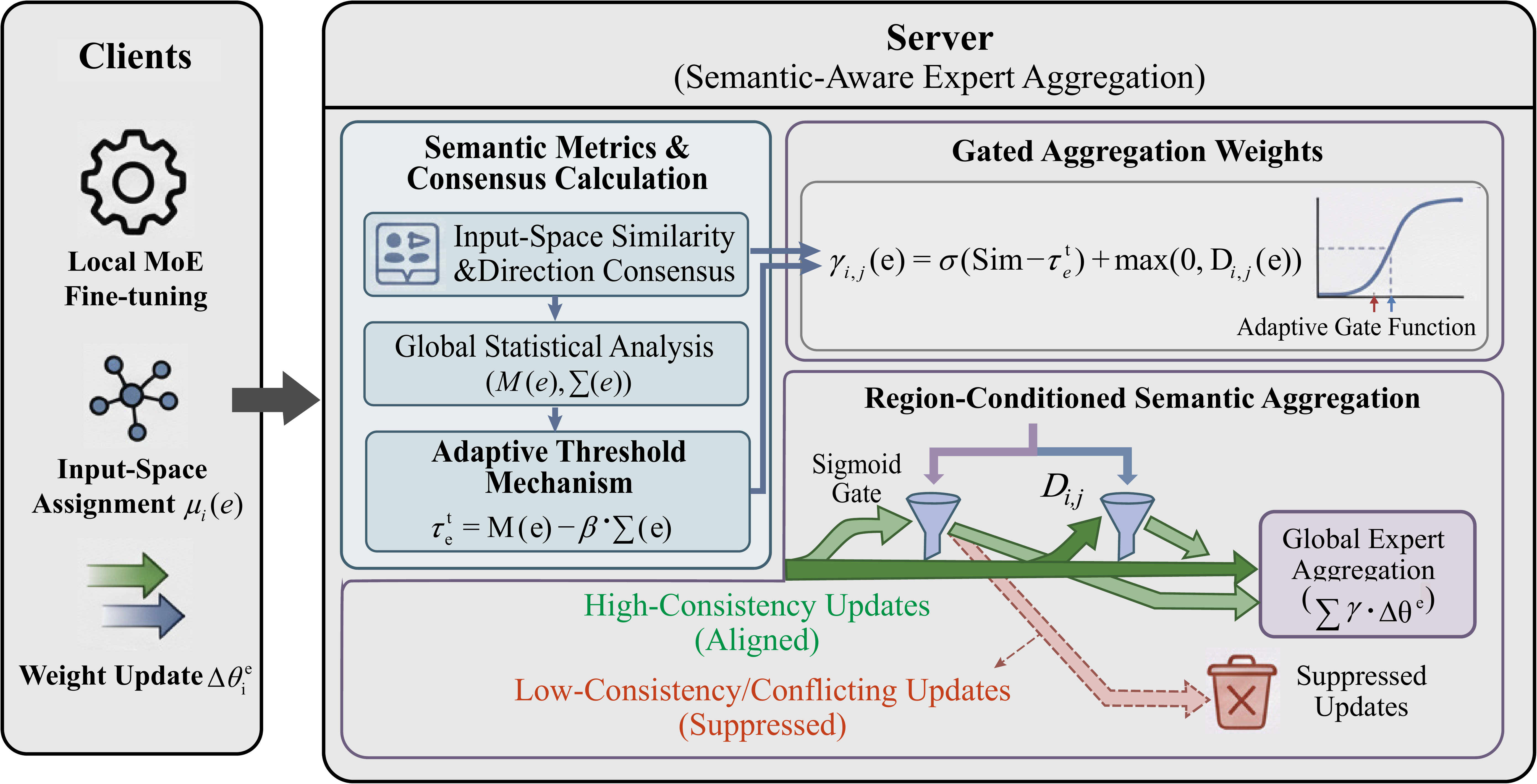}
\caption{The semantic-aware expert aggregation mechanism, which selectively aggregates experts based on semantic consistency, with adaptive thresholding to dynamically calibrate alignment sensitivity.}
\label{fig:expert_aggregation}
\vspace{-2ex}
\end{figure}

To preserve expert specialization in federated MoE fine-tuning, it is critical to address expert semantic misalignment caused by heterogeneous data distributions.
Unlike conventional methods that rely on the expert index for aggregation, \peerrev{we project experts into a functional space, providing a verifiable behavioral identity. 
Specifically, we characterize each expert's functional behavior using two semantic indicators derived from its input-space assignments and local weight updates.
By explicitly quantifying both the specific input regions an expert serves and the underlying feature–label correlations it has learned, these indicators effectively capture the expert's specialized role and learned knowledge for semantic alignment.}
As shown in Fig.~\ref{fig:expert_aggregation}, these expert semantics are then aligned through a region-conditioned aggregation strategy, which assigns gated aggregation weights to prioritize semantically consistent experts, thereby stabilizing expert semantics and preserving specialization.
These semantic indicators are detailed as follows.

\textbf{Input-space Assignment.}
\waitrev{In LLMs, the hidden representation space is highly structured, where spatial proximity inherently reflects semantic similarity.} As each expert receives a subset of hidden states corresponding to the inputs it processes, the average of these hidden states routed to an expert provides a concise representation of the semantic region (e.g., legal terminology or mathematical logic) that the expert specializes in for a given client. 
Large discrepancies between these representations across clients indicate that identically indexed experts are responsible for disparate semantic regions rather than a shared function.
Thus, during the local fine-tuning on client $i$, we define its input-space assignment $\mu_i(e)$ for expert $e$ as the average representation of the hidden states $h$ assigned to it by the gating network. Let $\mathcal{H}_i(e) = \{ h \mid \text{argmax } G(h) = e \}$ denote the set of hidden states routed to expert $e$. The input-space assignment is computed as
\begin{equation}
\mu_i(e) = \frac{1}{|\mathcal{H}_i(e)|} \sum_{h \in \mathcal{H}_i(e)} h.
\end{equation}

\textbf{Local Weight Update.}
While the input-space assignment $\mu_{i}(e)$ captures the semantic region that an expert specializes in, it does not fully characterize how the expert transforms those inputs into outputs.
\needrev{Even if two experts focus on similar input regions, differences in learned knowledge during local updates may induce divergent feature–label correlations.}
To capture this functional behavior, we use the local weight update $\Delta \theta$ as a semantic proxy,
since it reflects the expert’s optimization direction toward a specific input–output mapping. 
Consistent update directions across clients indicate a shared objective and functional agreement for same-indexed experts, whereas orthogonal or opposing updates indicate conflicts in learned knowledge \jnlrev{driven by divergent user behaviors on edge devices}.
After local fine-tuning, client $i$ computes the local weight update for each expert $e$ as $\Delta \theta_i^e = \theta_i^{e,new} - \theta_i^{e,old}$, and the direction consensus between client $i$ and $j$ is measured using the cosine similarity function $\text{Sim}(\cdot)$:
\vspace{-1ex}
\begin{equation}
D_{i,j}(e) = \text{Sim}(\Delta \theta_i^e, \Delta \theta_j^e) = \frac{\Delta \theta_i^e \cdot \Delta \theta_j^e}{\|\Delta \theta_i^e\| \|\Delta \theta_j^e\|}.
\end{equation}

To identify semantic alignment across clients and mitigate the adverse effects of misaligned experts, we design a region-conditioned gated aggregation mechanism that adaptively incorporates expert updates based on their semantic consistency.
Specifically, cosine similarity between input-space assignments acts as a primary filter to ensure that only experts focusing on similar input regions are compared, while the direction consensus measures the semantic alignment of their weight updates to ensure they are learning compatible transformations.
Based on these two criteria, the server assigns a region-conditioned aggregation weight $\gamma_{i,j}(e)$ to each expert $e$ for client pair $\{i,j\}$:
\begin{equation} \label{eq:semantic_aggregation_weight}
\gamma_{i,j}(e) = \mathbf{1}[\text{Sim}(\mu_i(e), \mu_j(e)) > \tau ] \cdot \max(0,D_{i,j}(e)),
\end{equation}
where $\tau$ represents the similarity threshold.
If the similarity between two experts' input regions falls below the threshold ($\text{Sim}(\mu_i(e), \mu_j(e)) \le \tau$), the gate closes, minimizing their mutual contribution to the global update.
For experts handling similar input regions, their contributions are further scaled by the direction consensus, with highly aligned updates receiving a high aggregation weight, while conflicting gradients are suppressed.

Finally, the global update for expert $e$ is computed via the weighted aggregation of local updates in training round $t$, which is expressed as
\vspace{-1ex}
\begin{equation} \label{eq:semantic_aggregation_expert}
\theta^e(t+1) = \theta^e(t) + \sum_{i=1}^N \frac{\sum_{j=1}^N {\gamma}_{i,j}(e)}{\sum_{i=1}^N \sum_{j=1}^N {\gamma}_{i,j}(e)} \Delta \theta_i^e.
\end{equation}

\subsubsection{\textbf{Selective Aggregation with Adaptive Threshold}}
\label{sec:dsn_expert_threshold}
Individual experts within an MoE architecture exhibit distinct behavior during fine-tuning~\cite{liu2023yoga, krizhevsky2012imagenet,  raghu2017svcca}, as some experts may rapidly converge to a globally consistent functional role, while others remain highly specialized to local data distributions.
These differences in specialization stabilization are further exacerbated by the non-IID data characteristic of federated environments \jnlrev{where mobile clients have highly skewed local
interactions}.
A static threshold $\tau$ in aggregation weight for expert semantic alignment fails to accommodate this diversity, where a large $\tau$ may discard valid collaborative knowledge, leading to \rev{under-aggregation}, whereas a small value incorporates semantically conflicting updates, resulting in semantic blurring and degraded expert specialization.
These limitations make it imperative to adapt the region-conditioned aggregation mechanism to the specialization discrepancy for each expert.

To enable robust and stable expert selective aggregation, we employ an adaptive thresholding mechanism to dynamically calibrate alignment sensitivity.
\peerrev{Since an expert's specialized role is inherently dictated by the data regions it processes, the statistic of semantic similarities in the input-space assignments measures how closely the local expert $e$ aligns with the \needrev{collective consensus} across clients. Thus, we use the statistical dispersion of expert semantics to calibrate aggregation weight threshold $\tau$.} 
Specifically, during global aggregation, the alignment sensitivity for expert $e$ is quantified by its mean semantic similarity $M(e)$ and dispersion $\Sigma(e)$, defined as follows:
\begin{equation}
\begin{aligned}
&M(e) = \frac{1}{N^2} \sum_{i=1}^N \sum_{j=1}^N S_{i,j}(e), \\
\Sigma(e) =& \sqrt{\frac{1}{N^2} \sum_{i=1}^N \sum_{j=1}^N {(S_{i,j}(e)-M(e))}^2},
\end{aligned}
\end{equation}
where $S_{i,j}(e)= \text{Sim}(\mu_i(e), \mu_j(e))$ denotes the set of pairwise similarities between client $i$ and $j$ for expert $e$.
Here, $M(e)$ represents the central reference of expert $e$’s semantic alignment across all clients, while $\Sigma(e)$ captures the dispersion of this alignment, reflecting the variance of consensus on the functional role of expert $e$. A high mean similarity with low dispersion indicates that expert $e$ has reached a stable and widely shared semantic role, whereas a low mean or high dispersion suggests ongoing specialization or semantic disagreement across clients.

To adapt the alignment sensitivity to this evolving semantic state, we define an expert-specific adaptive threshold $\tau_e^t$ as
\begin{equation} \label{eq:semantic_adaptive_thre}
\tau_e^t = M(e) - \beta \cdot \Sigma(e),
\end{equation}
where $\beta$ controls the tolerance to semantic variability. This formulation tightens the threshold when expert semantics are well aligned across clients and relaxes it when expert roles remain diverse, thereby enabling adaptive expert aggregation that enhances specialization without including semantically conflicting updates.

With the adaptive threshold $\tau_e^t$ for expert $e$, we refine the gated aggregation weight $\gamma_{i,j}(e)$ as follows:
\begin{equation} 
\gamma_{i,j}(e) = \sigma \left( S_{i,j}(e) - \tau_e^t \right) \cdot \max(0, D_{i,j}(e)).
\end{equation}

By integrating this adaptive thresholding into our region-conditioned aggregation strategy, we ensure that the global MoE model \rev{maintains sharp} expert specialization and stable convergence, particularly under non-IID data heterogeneity \jnlrev{intrinsic to mobile edge networks}.
We summarize the proposed aggregation alignment for federated MoE-based LLM fine-tuning in Algorithm 1.


\SetKwInOut{Input}{Require}
\SetKwProg{Fns}{Server Executes}{:}{}
\SetKwFunction{Fns}{Server Executes}
\SetKwProg{Fn}{}{:}{}
\SetKwFunction{Fa}{Auto Configuration}
\RestyleAlgo{ruled}
\LinesNumbered

\begin{algorithm}
\caption{Aggregation Alignment for MoE-based LLM Fine-tuning \jnlrev{in edge computing systems}}
\label{alg:name}
\setstretch{1.0}
\small
\textbf{Input:} Total rounds $T$, local epochs $E$, Local Dataset $D_c$\\
\textbf{Initialize:} Global expert weights $\{\theta^e(0)\}_{e \in \mathcal{E}}$ and global gating distribution $p_g^0(e)$.\\
\For{each round $t = 0$ to $T$}{
    \textbf{// Local Fine-tuning Phase \jnlrev{(on Edge Devices)}}\\
    \For{each client $i$ in parallel}{
        Download global experts $\theta^e(t)$ and global routing reference $p_g^t(e)$\;
        Calculate adaptive consistency weights $\alpha_i(e)$ via Eqn.~\eqref{eq:regular_alpha}\;
        Fine-tune local MoE-based LLM via $\mathcal{L}_{\text{total}} = \mathcal{L}_{\text{local}} + \lambda \mathcal{L}_{\text{reg}}$ (Eqn.~\eqref{eq:local_objective})\;
        Compute average routing distribution $\bar{p}_i(e)$ and routing decision margin $m_i(e)$ for gating distribution alignment\;
        Compute input-space assignment $\mu_i(e)$ and local updates $\Delta \theta_i^e$ for expert semantic aggregation\;
        Upload $\{ \overline{p}_i(e), \Delta \theta_i^e, \mu_i(e), m_i(e) \}$ to server\;
    }
    \textbf{// Global Aggregation Alignment Phase (on Server)}\\
    \For{each expert $e \in \mathcal{E}$}{
        Calculate routing consistency weight $\omega_i(e)$ with Eqn.~\eqref{eq:routing_consist_weight}\;
        Update global reference: $p_g^{t+1}(e)$\;
        Analyze statistics $M(e)$ and $\Sigma(e)$ to determine adaptive threshold $\tau_e^t$ via Eqn.~\eqref{eq:semantic_adaptive_thre}\;
        \For{each client pair $(i, j)$}{
            Compute input-space similarity and direction consensus $D_{i,j}(e)$\;
            Calculate region-conditioned weight $\gamma_{i,j}(e)$ using Eqn.~\eqref{eq:semantic_aggregation_weight}\;
        }
        Update global expert $\theta^e(t+1)$ with Eqn.~\eqref{eq:semantic_aggregation_expert}\;
    }
}
\textbf{Output:} Final global model $\theta_{\text{global}}^{T}$\\
\end{algorithm}

\begin{table*}[t]
\centering
\renewcommand{\arraystretch}{1.25}
  \begin{tabular}{c||c||cccc||cccc}
    \hline
    \multirow{2}{*}{Model} & \multirow{2}{*}{Method} & \multicolumn{4}{c||}{\textbf{IID distribution}} & \multicolumn{4}{c}{\textbf{non-IID distribution}} \\
    \cline{3-10}
    & & AGNews & PIQA & HellaSwag & MMLU & AGNews & PIQA & HellaSwag & MMLU \\
    \hline
    \multirow{5}{1.1cm}{Switch-base-16}
    & \texttt{FedAvg}  & 0.9263 & 0.7421 & 0.7172 & 0.4641 & 0.7740 & 0.6792 & 0.5206 & 0.3008 \\
    & \texttt{FedProx} & 0.9290 & 0.7473 & 0.7204 & 0.4707 & 0.7872 & 0.6878 & 0.5288 & 0.3176 \\
    & \texttt{PFL-MoE}  & 0.9318 & 0.7596 & 0.7314 & 0.4837 & 0.7994 & 0.7008 & 0.5476 & 0.3325 \\
    & \texttt{FedMoE} & 0.9339 & 0.7681 & 0.7446 & 0.4812 & 0.8137 & 0.7126 & 0.5642 & 0.3506 \\
    & \texttt{\name}  & \textbf{0.9424} & \textbf{0.7822} & \textbf{0.7531} & \textbf{0.5131} & \textbf{0.8522} & \textbf{0.7325} & \textbf{0.5847} & \textbf{0.3845} \\
    \hline
    \multirow{5}{1.5cm}{DeepSeek-MoE-16B}
    & \texttt{FedAvg}  & 0.9206 & 0.7531 & 0.7356 & 0.3960 & 0.7696 & 0.6824 & 0.5105 & 0.2543 \\
    & \texttt{FedProx} & 0.9244 & 0.7602 & 0.7423 & 0.4042 & 0.7816 & 0.6918 & 0.5179 & 0.2589 \\
    & \texttt{PFL-MoE}  & 0.9272 & 0.7877 & 0.7508 & 0.4125 & 0.7924 & 0.7136 & 0.5344 & 0.2658 \\
    & \texttt{FedMoE} & 0.9289 & 0.7928 & 0.7611 & 0.4271 & 0.8112 & 0.7214 & 0.5480 & 0.2805 \\
    & \texttt{\name}  & \textbf{0.9402} & \textbf{0.8115} & \textbf{0.7782} & \textbf{0.4424} & \textbf{0.8518} & \textbf{0.7410} & \textbf{0.5711} & \textbf{0.3336} \\
    \hline
  \end{tabular}
\vspace{-0.5ex}
\caption{The test accuracy across benchmarks on four datasets with Switch-base-16 and DeepSeek-MoE-16B models under IID and non-IID data distribution ($\alpha=0.1$).}
\vspace{-3ex}
\label{tab:overall_performance}
\end{table*}

\section{Simulation Setup} \label{sec:simulation_simu_setup}
In this section, we present the detailed experimental setup of \name for heterogeneous MoE-based LLM fine-tuning, which is evaluated against several baselines using carefully selected hyper-parameters to ensure a fair comparison.

\subsubsection{\textbf{Models}}
For our experiments, we employ two representative MoE-based LLMs, Switch Transformer~\cite{fedus2022switch} and DeepSeek-MoE-16B~\cite{dai2024deepseekmoe}. 
The Switch Transformer represents a classic scalable MoE design, utilizing a Top-1 routing protocol that directs each token to one of 16 experts per layer. This mechanism allows the model to scale up to 1.6 trillion parameters while maintaining constant computational costs per token. DeepSeek-MoE-16B employs a more sophisticated routing strategy (Top-2 out of 64 experts), which demonstrates competitive performance with its dense model LLaMA2-7B. 

\subsubsection{\textbf{Datasets}}
To evaluate model performance, we report test accuracy for \rev{converged global model} on widely used datasets, covering tasks from basic semantic classification to complex knowledge reasoning:
(1)~AGNews~\cite{zhang2015character}: A multi-class news topic classification dataset containing four categories (World, Sports, Business, and Sci/Tech), commonly used to evaluate text classification capabilities.
(2)~PIQA~\cite{bisk2020piqa}: A binary-choice natural language understanding benchmark for commonsense reasoning, where models must choose the more plausible solution to a given goal involving everyday physical interactions.
(3)~HellaSwag~\cite{zellers2019hellaswag}: A four-way multiple-choice dataset designed to evaluate a model's ability to predict the most plausible solution of a given context. It employs adversarial filtering to ensure that models rely on deep reasoning rather than superficial statistical patterns. 
(4)~MMLU~\cite{hendrycks2020measuring}: A multi-task benchmark that
evaluates general knowledge and reasoning ability across 57 academic subjects, providing a comprehensive assessment of a model’s understanding capabilities covering diverse academic and professional domains.

\subsubsection{\textbf{Baselines}}
To investigate the advantages of our \name framework, we compare it with the following representative baselines:
\begin{itemize}
  \item \textbf{FedAvg~\cite{mcmahan17a}} is the classical federated optimization method that aggregates local models by directly averaging their parameters across clients.
  \item \textbf{FedProx~\cite{luo2023optimization}} extends FedAvg by introducing a proximal regularization term in the local objective, which constrains local updates to remain close to the global model and improves training stability under data heterogeneity.
  \item \textbf{PFL-MoE~\cite{guo2021pfl}} adopts a MoE architecture to enable personalized federated learning, where different experts specialize in distinct client distributions, enabling client-specific expert selection from globally shared knowledge.
  \item \textbf{FedMoE~\cite{mei2024fedmoe}} constructs an optimal sub-MoE for each client while jointly aggregating experts and the gating network across clients, striking the balance between efficiency and performance.
\end{itemize}

\begin{figure}[t!]
\vspace{-2.5ex}
\centering
\includegraphics[width=0.95\linewidth]{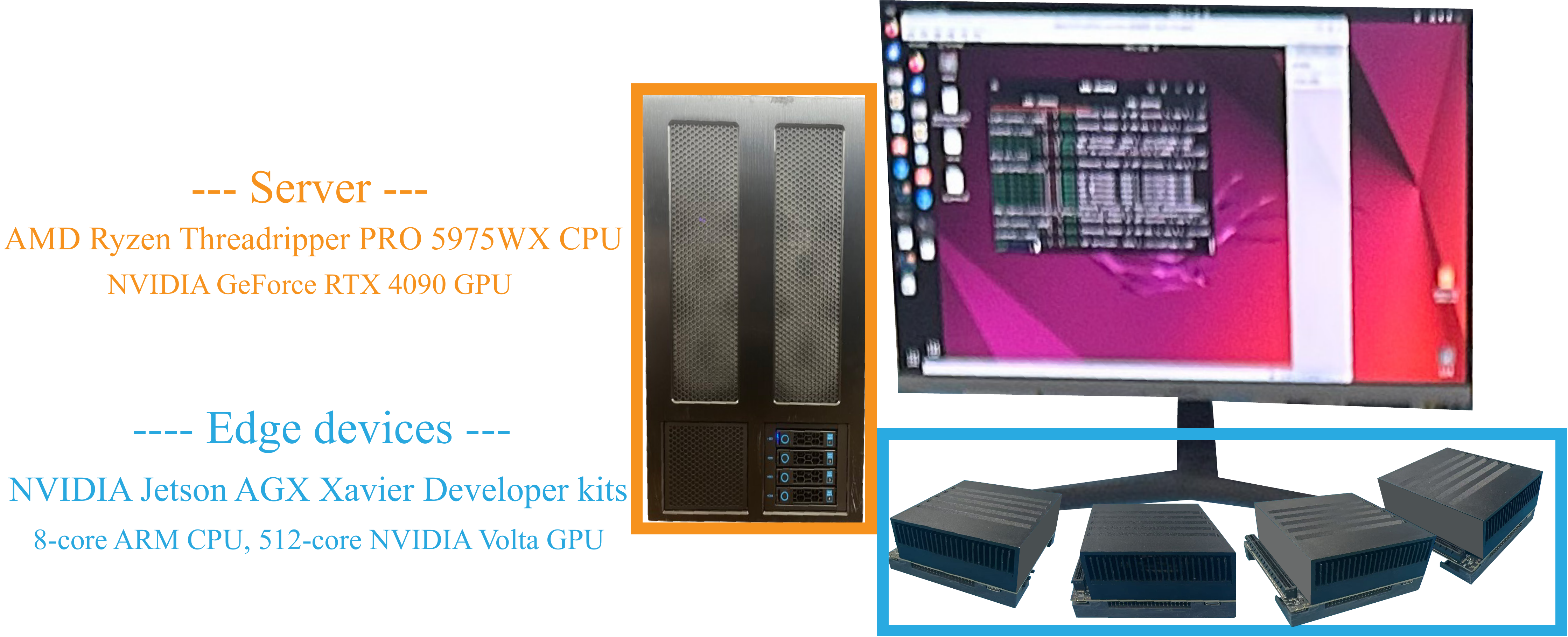}
\caption{The testbed for our \name framework.}
\label{fig:implementation}
\vspace{-2ex}
\end{figure}

\subsubsection{\textbf{Hyper-parameters}} 
In the experiments, we implement \name framework for MoE-based LLM instruction fine-tuning using Switch-base-16 and DeepSeek-MoE-16B as backbone models. Here, we employ 8-rank LoRA to enable efficient fine-tuning of the large-scale DeepSeek-MoE-16B model.
As illustrated in Fig.~\ref{fig:implementation}, the distributed system consists of a central server equipped with NVIDIA GeForce RTX 4090 GPU and $N=10$ edge devices running on NVIDIA Jetson AGX Xavier Developer kits. The server completes the aggregation after receiving updates from all participating clients.
Each client performs local fine-tuning with a learning rate of $1\times10^{-4}$ for one epoch per communication round, and the total number of communication rounds is set to 25. The key hyper-parameters in \name are set to $\lambda = 0.1$ and $\eta = 0.1$.
To simulate realistic data heterogeneity in federated learning, we adopt the commonly used Dirichlet-based data partitioning scheme~\cite{guo2021pfl, hu2025fft, huang2025fedsr}. Specifically, the client-specific label distributions are sampled from a Dirichlet distribution with concentration parameter $\alpha = 0.1$, and the training data are allocated accordingly. This makes each client exposed to a skewed subset of classes, thus creating a challenging non-IID setting for evaluating aggregation alignment.

\begin{figure*}[t]
\vspace{-1ex}
\centering
\subfloat[AGNews \label{fig:local_acc_agnews}]{
\includegraphics[width=0.24\linewidth]{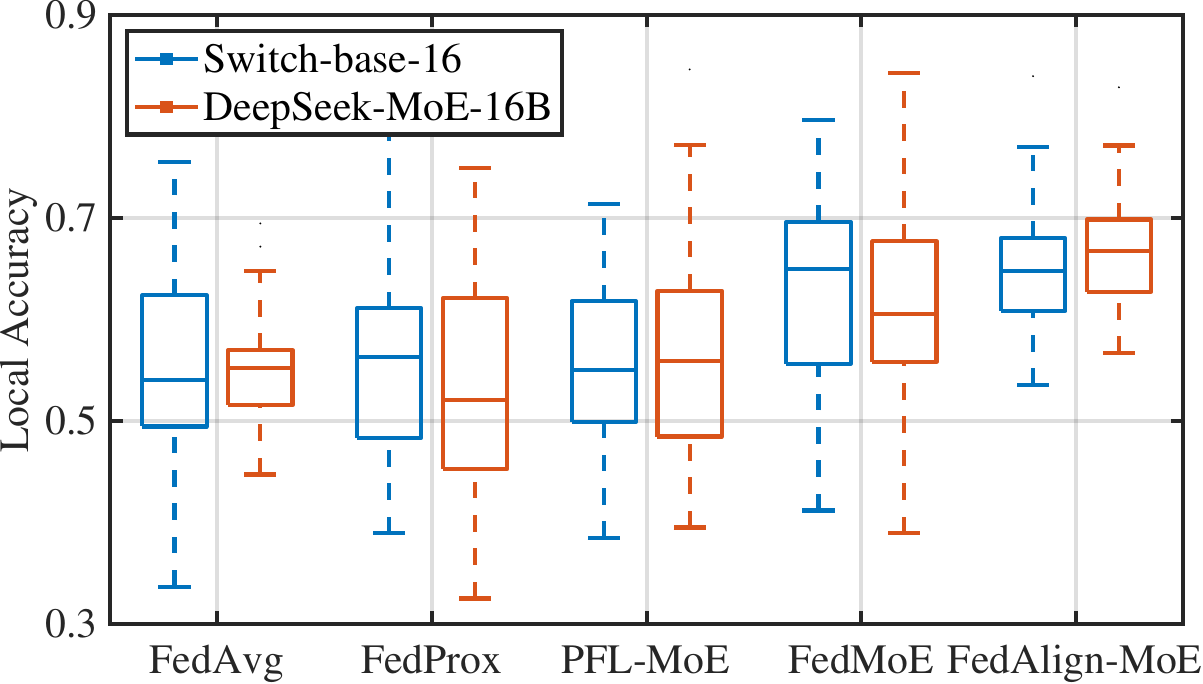}
}
\subfloat[PIQA \label{fig:local_acc_piqa}]{
\includegraphics[width=0.24\linewidth]{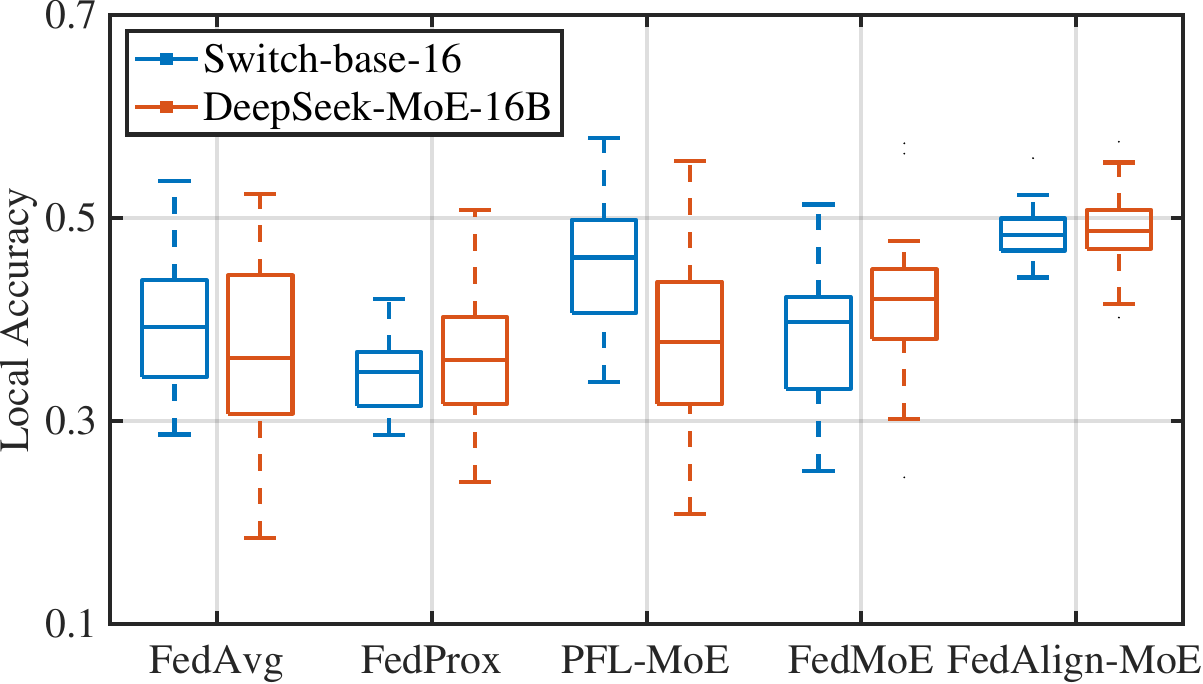}
}
\subfloat[HellaSwag \label{fig:local_acc_hellaswag}]{
\includegraphics[width=0.24\linewidth]{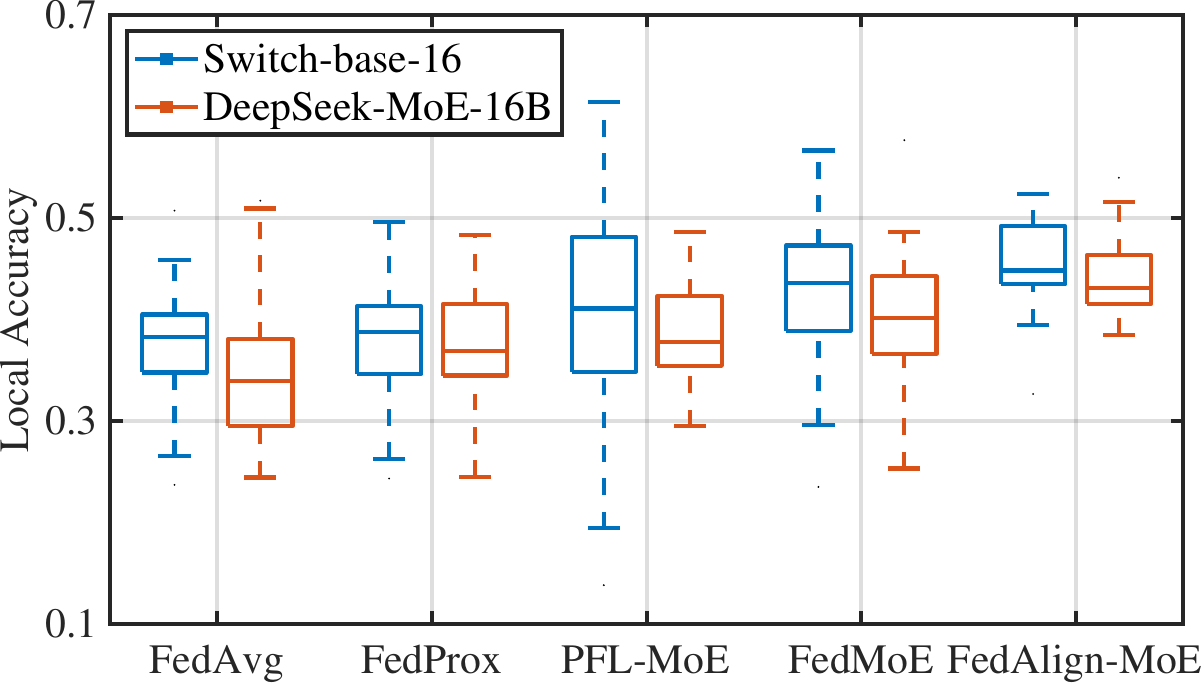}
}
\subfloat[MMLU \label{fig:local_acc_mmlu}]{
\includegraphics[width=0.24\linewidth]{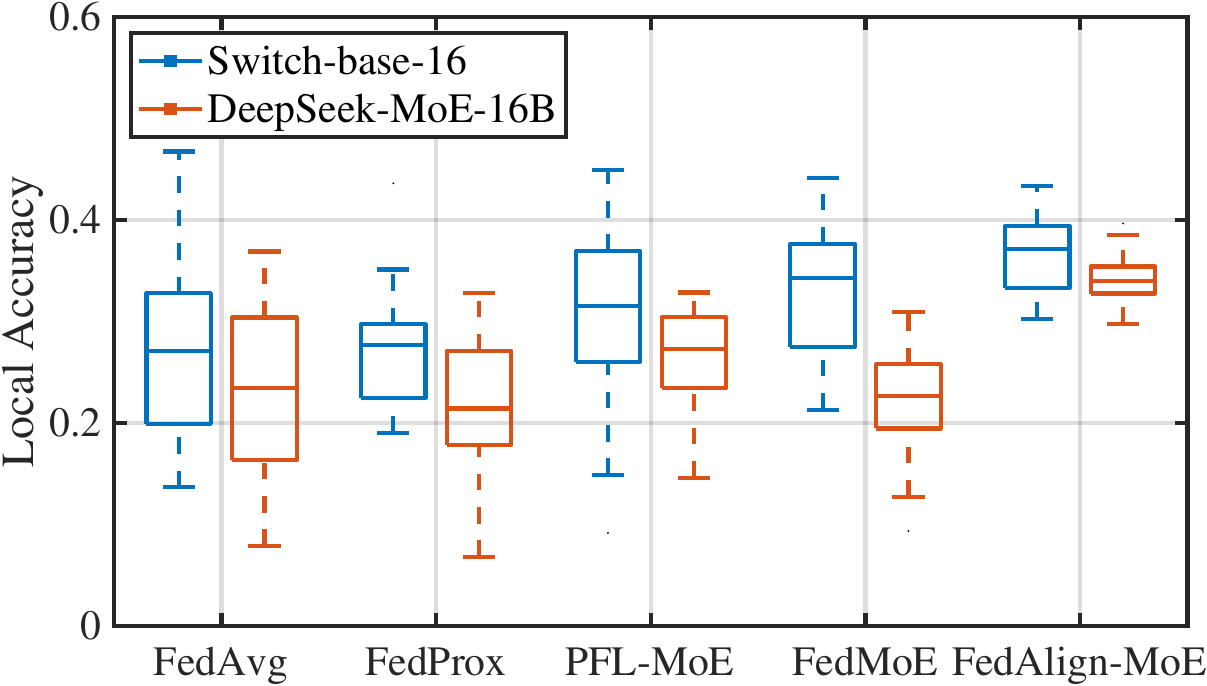}
}
\vspace{-1ex}
\caption{The average accuracy for local fine-tuning on four datasets under non-IID data distributions.} 
\label{fig:local_accuracy}
\vspace{-2ex}
\end{figure*}

\section{{Performance Evaluation}}\label{sec:simulation}
In this section, we evaluate the effectiveness of \name against various benchmarks under data heterogeneity. We further investigate the robustness of the proposed framework under varying data distributions and hyper-parameter settings.
\rev{The contributions of each key components in \name are also analyzed to illustrate their individual roles in the proposed framework.}

\vspace{-1ex}
\subsection{The Overall Performance}
\subsubsection{\textbf{The converged accuracy}}
Table~\ref{tab:overall_performance} presents the test accuracy of \name and other benchmarks on four datasets under IID and non-IID data distributions. 
In the IID setting, all FL methods achieve comparable performance as aligned gating preferences allow experts to learn consistent semantic functions across clients. \name outperforms other baselines across diverse datasets, indicating that consistency-based gating distribution alignment refines global routing behaviors and enhances expert specialization even under uniform distributions.
In the non-IID setting, conventional approaches suffer significant performance degradation due to \rev{the failure of direct parameter aggregation on routing preference preservation and expert semantic alignment.}
Although FedProx constrains local updates to the global model via proximal regularization, the neglect of client-specific routing preferences in global model still leads to suboptimal expert selection and degraded performance.
Compared to MoE-based LLM fine-tuning, such as FedMoE and PFL-MoE, despite performing expert-level aggregation or weighted combination of local and global models to balance personalization and generalization, the gating preference alignment and expert semantic aggregation are still unexplored.
In contrast, by explicitly aligning gating distributions and aggregating semantically aligned expert updates, \name consistently outperforms all baselines in all datasets, achieving an accuracy improvement of \needrev{4\% over FedMoE and 7\% over FedAvg}.
\waitrev{To further demonstrate the robustness and stability of \name, we report the mean and deviation of local fine-tuning accuracy in Fig.~\ref{fig:local_accuracy}, highlighting its ability to achieve aligned client-specific routing preference and stabilized expert specification.}


\begin{figure}[t]
\vspace{-3ex}
\centering
\subfloat[AGNews \label{fig:convergence_agnews}]{
\includegraphics[width=0.49\linewidth]{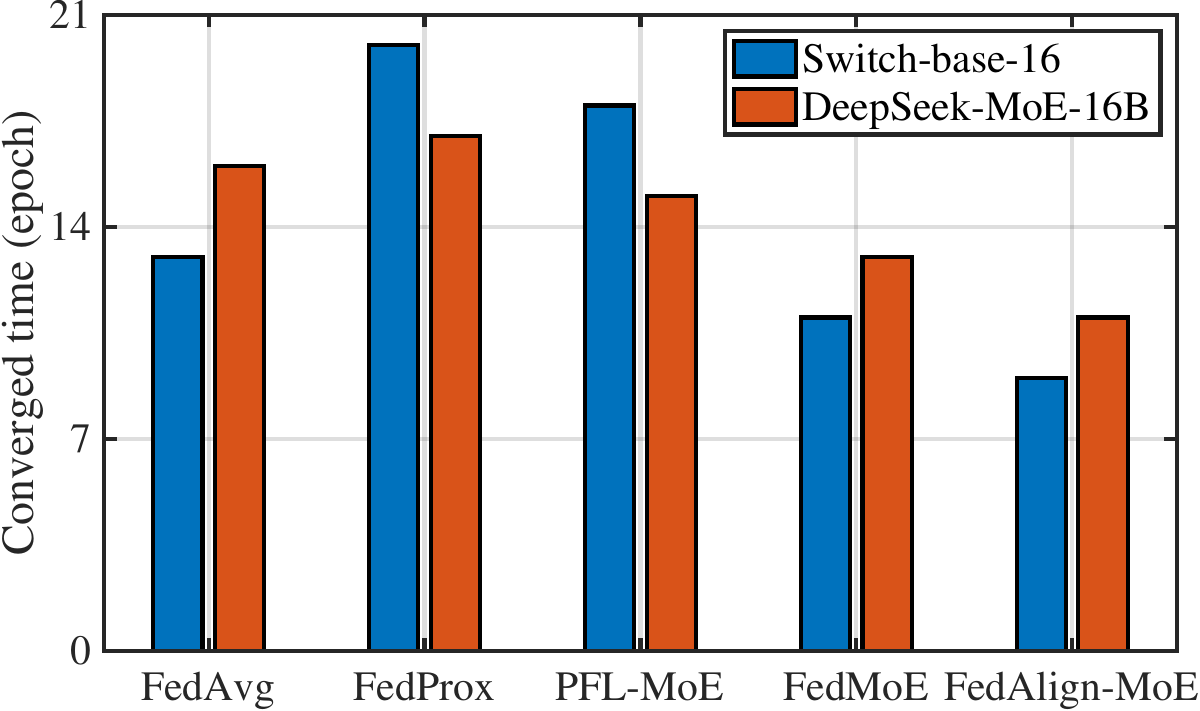}
}
\subfloat[MMLU \label{fig:convergence_mmlu}]{
\includegraphics[width=0.49\linewidth]{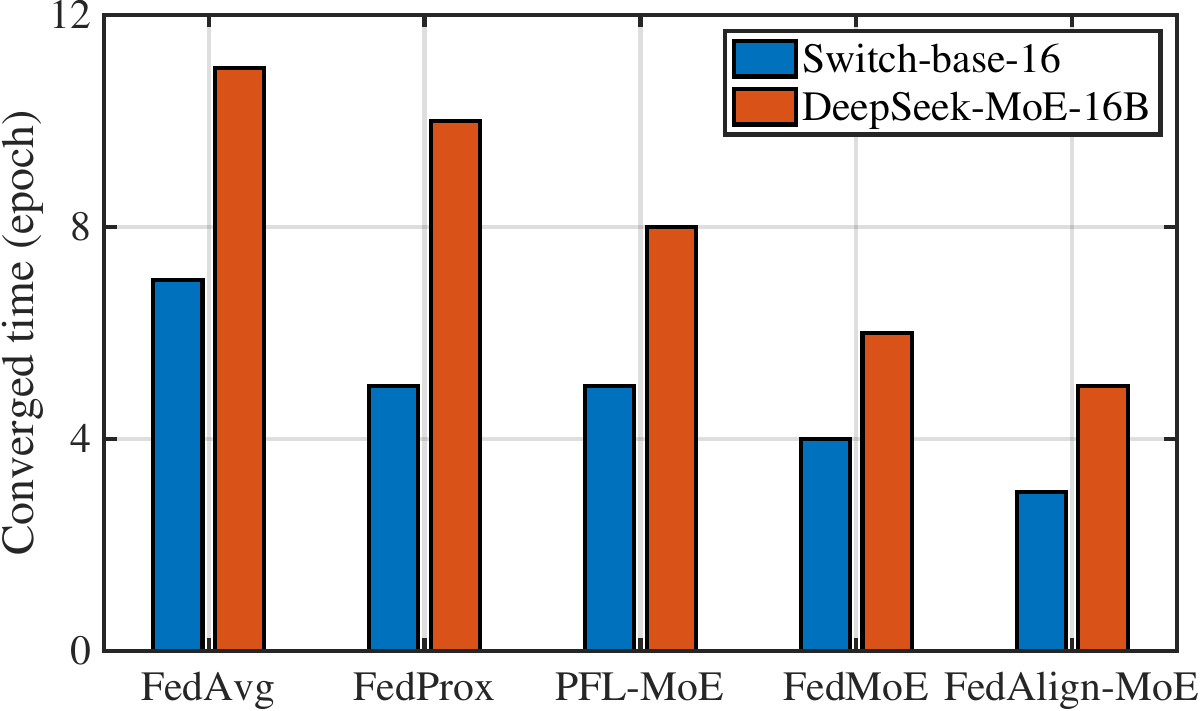}
}
\vspace{-1ex}
\caption{The convergence time across benchmarks on AGNews and MMLU datasets under non-IID data distributions.}
\label{fig:convergence_time}
\vspace{-2ex}
\end{figure}

\subsubsection{\textbf{The Convergence Performance}}
Fig.~\ref{fig:convergence_time} compares the convergence time of \name and other four benchmarks on AGNews (classification) and MMLU (reasoning) tasks using Switch-base-16 and DeepSeek-MoE-16B models \waitrev{under non-IID data distributions}. 
We observe that \name consistently achieves fastest convergence, yielding over \needrev{x1.3} and \needrev{x1.7} speedups on Switch-base-16 and DeepSeek-MoE-16B models, respectively.
This improvement primarily stems from its semantic-aware expert aggregation, which explicitly aligns expert semantics across clients, thereby stabilizing the semantic roles of same-indexed experts across clients and significantly accelerating convergence under heterogeneous data distributions.
Although FedMoE supports client-specific expert selection while FedProx and PFL-MoE enhance local adaptation, they still fail to enforce semantic consistency of same-indexed experts. As a result, experts with different semantic roles are aggregated together, introducing conflicting optimization directions and resulting in unstable convergence.

\begin{figure}[t]
\vspace{-3ex}
\centering
\subfloat[Switch-base-16]{
\includegraphics[width=0.49\linewidth]{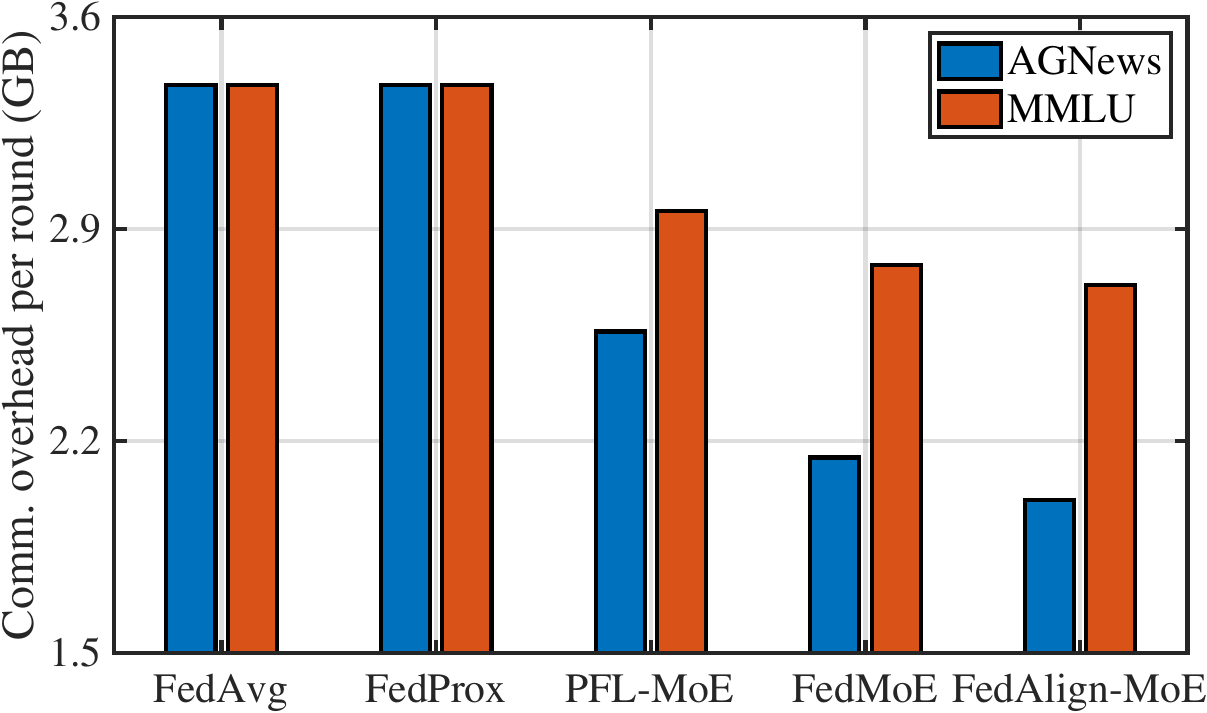}
}
\subfloat[DeepSeek-MoE-16B]{
\includegraphics[width=0.49\linewidth]{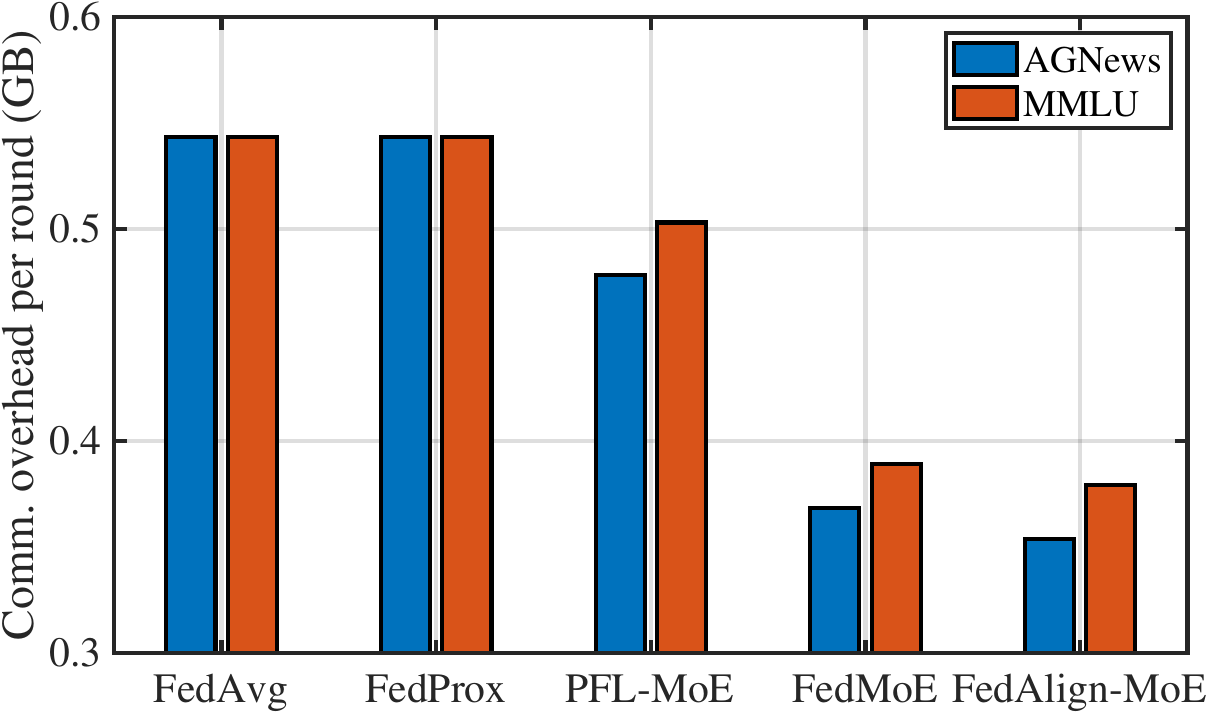}
}
\vspace{-1ex}
\caption{The comparison of average communication overhead on AGNews and MMLU datasets.}
\label{fig:communication_overhead}
\vspace{-4ex}
\end{figure}


\begin{table*}[t]
\centering
\renewcommand{\arraystretch}{1.25}
  \begin{tabular}{c|c|ccc|ccc}
    \hline
    \multirow{2}{*}{Model} & \multirow{2}{*}{Method} & \multicolumn{3}{c|}{\textbf{AGNews}} & \multicolumn{3}{c}{\textbf{MMLU}} \\
    \cline{3-8}
    & & $\alpha=0.1$ & $\alpha=0.5$ & $\alpha=1.0$ & $\alpha=0.1$ & $\alpha=0.5$ & $\alpha=1.0$ \\
    \hline
    \multirow{5}{1.1cm}{Switch-base-16}
    & \texttt{FedAvg} & 0.7740 & 0.8866 & 0.8978 & 0.3008 & 0.3482 & 0.4012 \\ 
    & \texttt{FedProx} & 0.7872 & 0.9006 & 0.9048 & 0.3176 & 0.3615 & 0.4148 \\ 
    & \texttt{PFL-MoE} & 0.7994 & 0.9041 & 0.9109 & 0.3325 & 0.3789 & 0.4305 \\ 
    & \texttt{FedMoE} & 0.8137 & 0.9103 & 0.9164 & 0.3506 & 0.3921 & 0.4463 \\ 
    & \texttt{\name} & \textbf{0.8522} & \textbf{0.9312} & \textbf{0.9355} & \textbf{0.3845} & \textbf{0.4250} & \textbf{0.4766} \\
    \hline
    \multirow{5}{1.5cm}{DeepSeek-MoE-16B}
    & \texttt{FedAvg} & 0.7696 & 0.8810 & 0.8965 & 0.2543 & 0.3060 & 0.3728 \\
    & \texttt{FedProx} & 0.7816 & 0.8963 & 0.9002 & 0.2589 & 0.3178 & 0.3984 \\
    & \texttt{PFL-MoE} & 0.7924 & 0.9017 & 0.9100 & 0.2658 & 0.3275 & 0.4005 \\
    & \texttt{FedMoE} & 0.8112 & 0.9096 & 0.9122 & 0.2805 & 0.3508 & 0.4146 \\
    & \texttt{\name} & \textbf{0.8518} & \textbf{0.9281} & \textbf{0.9338} & \textbf{0.3336} & \textbf{0.3953} & \textbf{0.4331} \\
    \hline
  \end{tabular}
\vspace{-0.5ex}
\caption{The test accuracy on the AGNews and MMLU datasets under non-IID data distributions using Switch-base-16 and DeepSeek-MoE-16B models.}
\label{tab:data_heterigeneity}
\vspace{-6ex}
\end{table*}

\subsubsection{\textbf{Deployment Efficiency}}
\begin{figure}[t]
\centering
\subfloat[Switch-base-16]{
\includegraphics[width=0.49\linewidth]{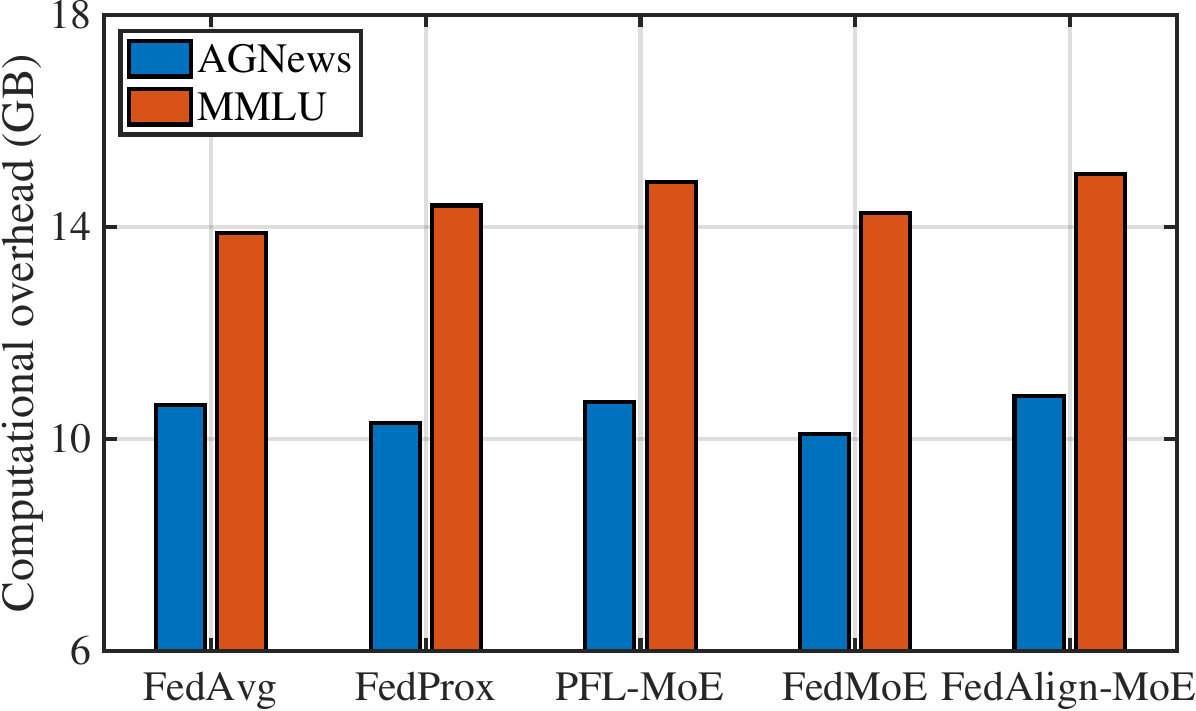}
}
\subfloat[DeepSeek-MoE-16B]{
\includegraphics[width=0.49\linewidth]{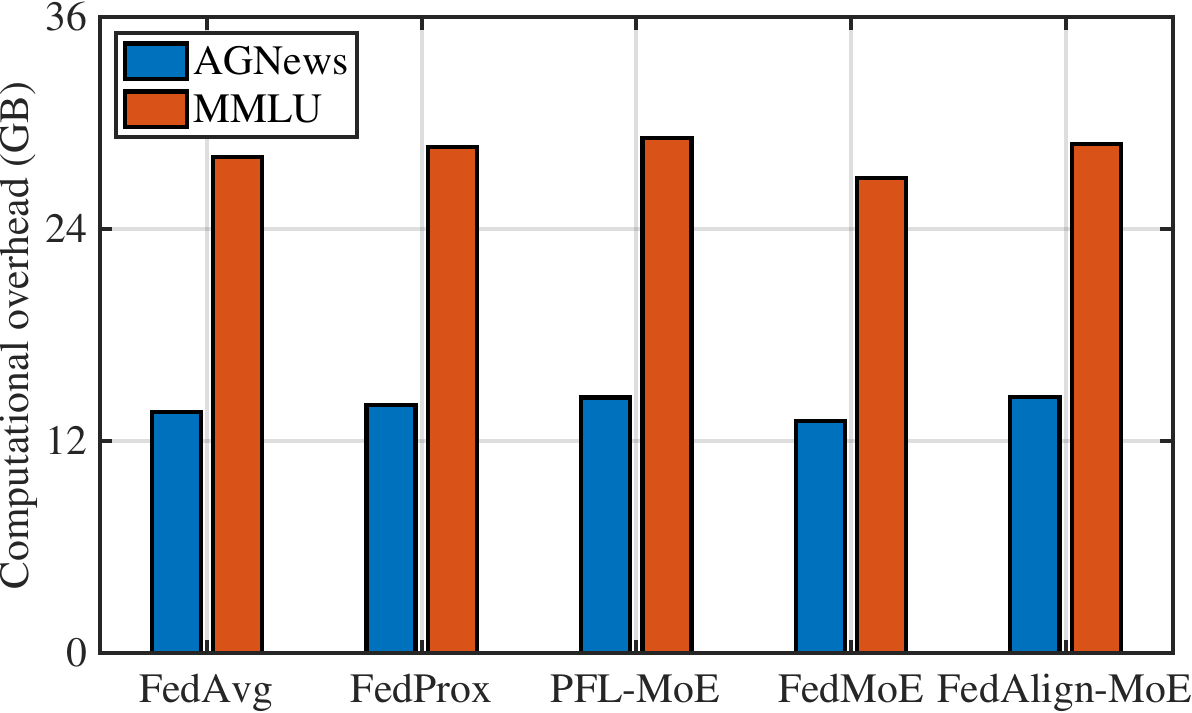}
}
\vspace{-1ex}
\caption{The comparison of average GPU memory footprint across clients on AGNews and MMLU datasets.}
\label{fig:computational_overhead}
\vspace{-2ex}
\end{figure}

\jnlrev{Fig.~\ref{fig:communication_overhead} illustrates the communication overhead as the average volume of uploaded data from each client for Switch-base-16 and DeepSeek-MoE-16B models.
Traditional baselines like FedAvg and FedProx incur severe communication bottlenecks, as they necessitate uploading the full model parameters, including all experts and gating networks.
In contrast, sparse activation frameworks such as FedMoE and pFL-MoE mitigate this payload by uploading only the activated experts and gating networks. Compared to these efficient baselines, the additional routing statistics ($m_i(e)$) and semantic metrics ($\mu_i(e)$) in \name introduce a negligible communication overhead of only 0.564 MB for Switch-base-16 and 5.07 MB for DeepSeek-MoE-16B. This accounts for less than 1.4\% of the model parameter transmission cost and even less than the cost of uploading gating parameters, demonstrating the communication-efficiency of \name for edge networks.
We further report the client-side computational overhead of each method using the average GPU memory footprint across clients during local fine-tuning in Fig.~\ref{fig:computational_overhead}. Compared to the local cost of standard FedAvg, the additional routing statistics computation and semantic metric extraction in \name introduce less than 6\% of the total GPU memory footprint, confirming that \name is computationally efficient for mobile edge deployments.
}

\subsubsection{\textbf{Scalability}} 
\begin{figure}[t]
\centering
\subfloat[AGNews \label{fig:scalable_llama2}]{
\includegraphics[width=0.49\linewidth]{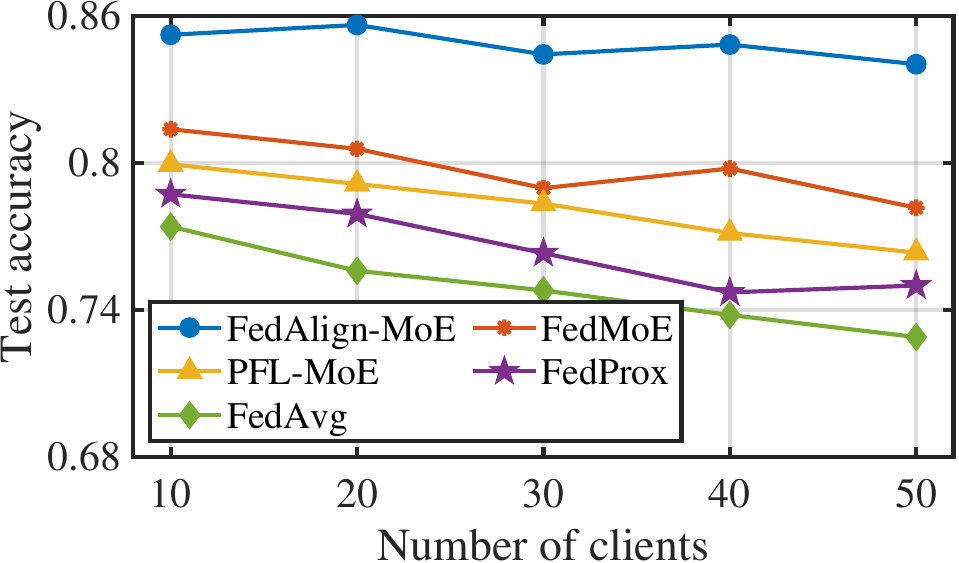}}
\subfloat[MMLU \label{fig:scalable_gpt2}]{
\includegraphics[width=0.49\linewidth]{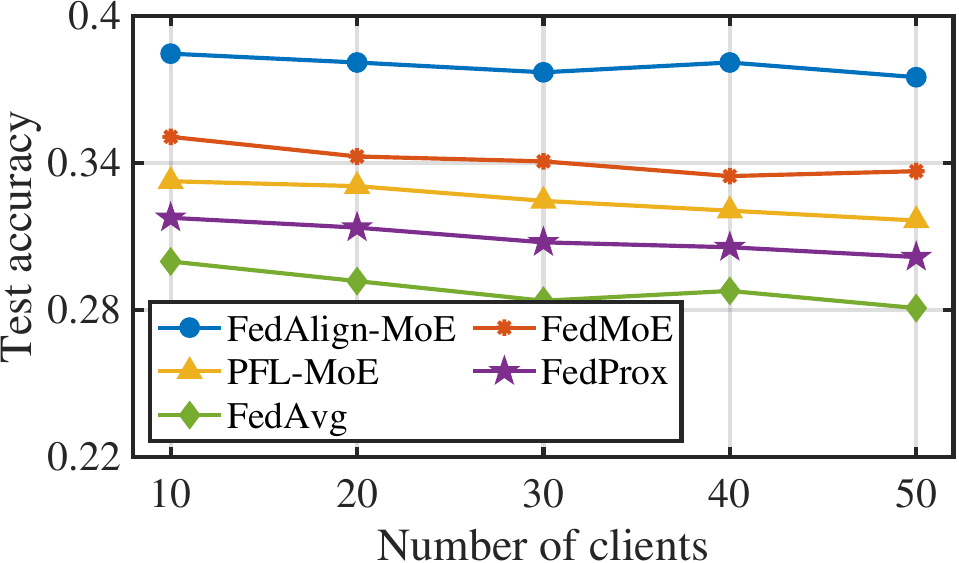}}
\vspace{-1ex}
\caption{The scalability of \name and other baselines on AGNews and MMLU datasets under non-IID local data.}
\label{fig:scalability}
\vspace{-2ex}
\end{figure}

Fig.~\ref{fig:scalability} demonstrates the scalability of \name with other baselines \needrev{on the Switch-base-16 model} as the client population increases from 10 to 50 under non-IID data distributions.
It can be observed that as the number of clients increases, the test accuracy of all methods exhibits a slight yet stable decrease, as a larger client population introduces more diverse and non-IID distributions exacerbates expert semantic divergence. 
However, \name consistently achieves superior performance and robust scalability among all benchmarks, underscoring its adaptability to practical large-scale FL deployments.
This robustness stems from the explicit alignment of both routing behaviors and expert semantic roles across clients during global aggregation, rather than relying solely on direct parameter averaging or sparsity-based expert selection, both of which become increasingly vulnerable to routing misalignment and expert semantic divergence as more heterogeneous clients participate.

\vspace{-1ex}
\subsection{Micro-benchmarking}
\subsubsection{\textbf{Robustness to non-IID Data Distribution}}
Table.~\ref{tab:data_heterigeneity} reports the impact of heterogeneous data distributions for \name and other baselines on converged test accuracy using AGNews and MMLU datasets.
While all baseline methods exhibit inevitable accuracy
degradation as $\alpha$ decreases from $1.0$ to $0.1$, \name consistently preserves over \needrev{90\% and 75\%} of its full-performance accuracy on Switch-base-16 and DeepSeek-MoE-16B models, demonstrating remarkable robustness to severe data heterogeneity.
By aligning gating behaviors with local distribution regularization, \name ensures that expert selection remains adaptive to local data characteristics without compromising global routing consistency.
By modeling the functional roles of experts and prioritizing semantically consistent updates, \name further preserves expert specialization across local data.
Conversely, the significant performance drop in baselines such as FedAvg and FedMoE is primarily due to their reliance on index-based expert aggregation, which ignores the fact that local data distributions may cover fundamentally different feature regions in highly non-IID scenarios, causing severe semantic divergence among identically indexed experts.
Although FedProx and PFL-MoE enhance local adaptation through regularization or personalization, they fail to account for the misalignment of local gating behaviors, resulting in ineffective expert selection and further limiting their robustness.

\begin{figure}[t]
\vspace{-2ex}
\centering
\subfloat[AGNews]{
\includegraphics[width=0.49\linewidth]{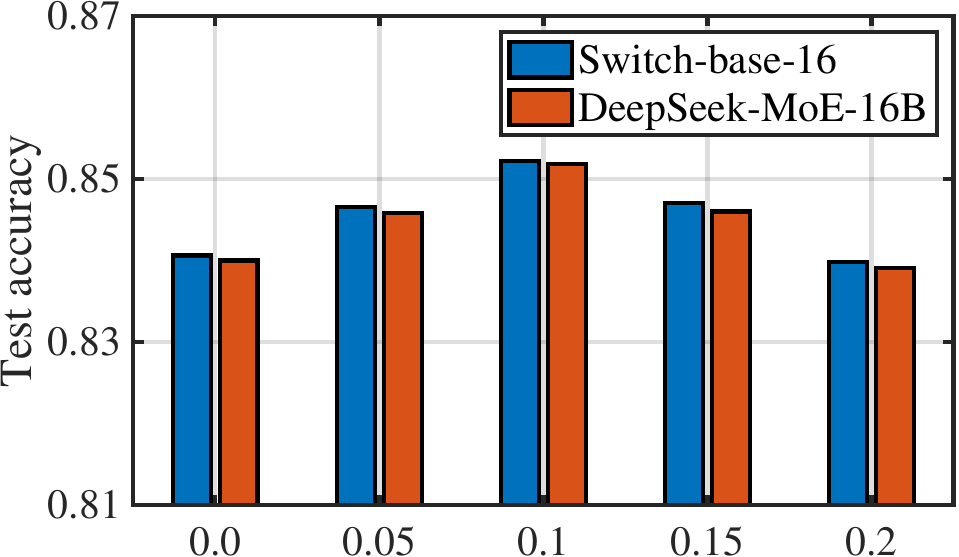}
}
\subfloat[MMLU]{
\includegraphics[width=0.49\linewidth]{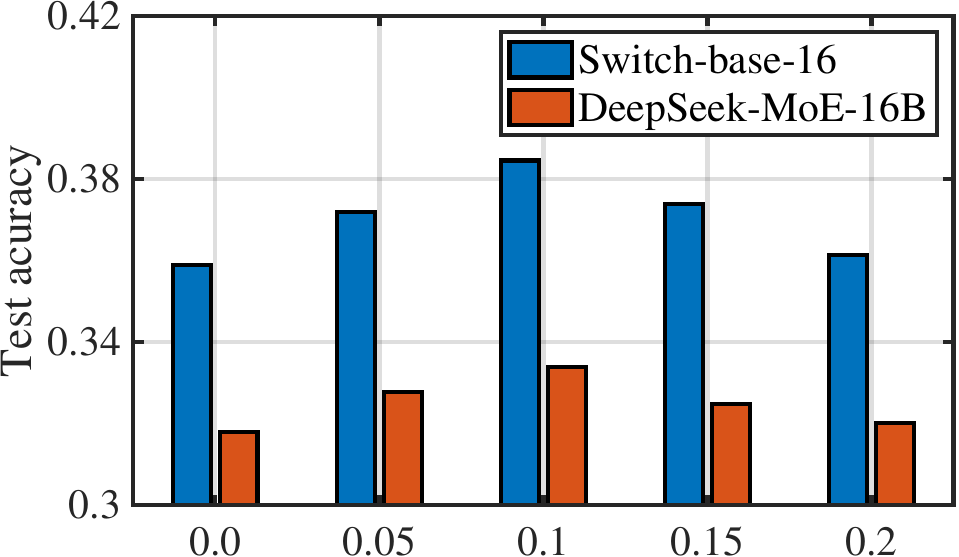}
}
\vspace{-1ex}
\caption{Performance of \name with varying expert overlap threshold $\eta$ on AGNews and MMLU datasets.}
\label{fig:hyperparameter_eta}
\vspace{-2ex}
\end{figure}

\begin{figure}[t]
\vspace{-1.5ex}
\centering
\subfloat[AGNews]{
\includegraphics[width=0.49\linewidth]{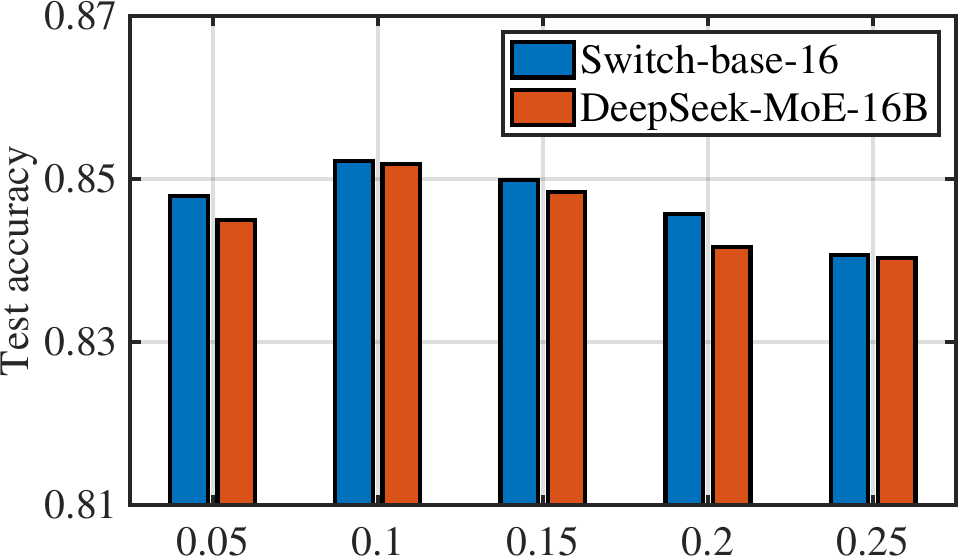}
}
\subfloat[MMLU]{
\includegraphics[width=0.49\linewidth]{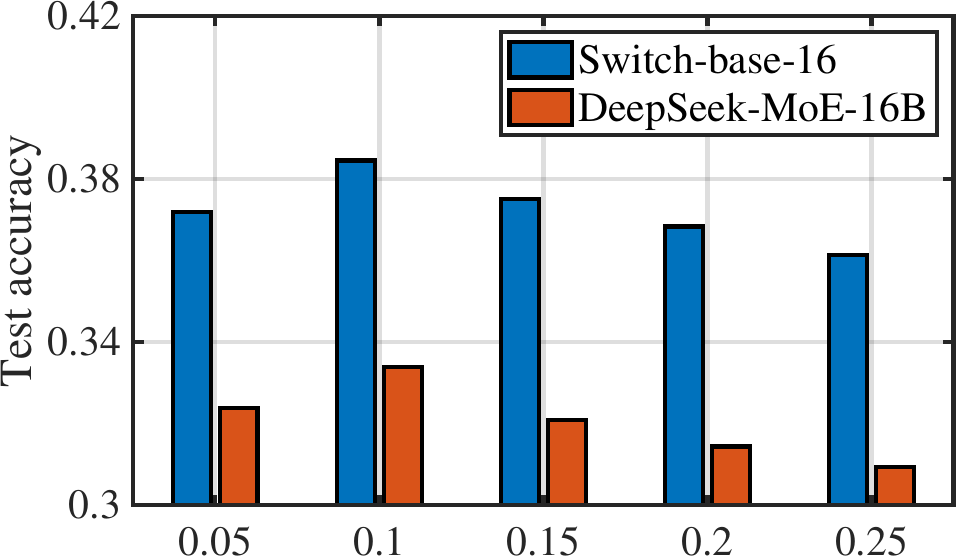}
}
\vspace{-1ex}
\caption{Performance of \name with different  balancing coefficient $\lambda$ on AGNews and MMLU datasets.}
\label{fig:hyperparameter_lambda}
\vspace{-2ex}
\end{figure}

\subsubsection{\textbf{Sensitivity Analysis}}
Fig.~\ref{fig:hyperparameter_eta} illustrates the impact of varying expert overlap threshold $\eta$ on test accuracy over the AGNews and MMLU datasets, where $\eta$ controls the sensitivity of expert alignment during client-side distribution regularization.
When $\eta$ is too low, most routing decisions are regarded as globally consistent, resulting in strong regularization that forces alignment even on experts that have specialized in local data features and suppresses discriminative local preferences. 
In contrast, a large $\eta$ weakens the alignment constraint by assigning small weights to most experts, leading to reduced cross-client consistency and suboptimal accuracy.
Moreover, Fig.~\ref{fig:hyperparameter_lambda} presents the impact of varying the coefficient $\lambda$, which balances task adaptation and gating consistency in the local objective. 
When $\lambda$ is close to zero, the regularization loss becomes negligible, causing gating networks to \rev{drift independently across clients} and leading to severe routing misalignment after aggregation. 
As $\lambda$ increases, performance improves as stronger regularization encourages consistent routing behaviors and stabilizes expert selection across clients. 
However, large $\lambda$ overly constrains local optimization, preventing gating networks from adapting to client-specific data distributions and thus degrading task performance.

\begin{figure}[t]
\vspace{-2ex}
\centering
\subfloat[AGNews]{
\includegraphics[width=0.49\linewidth]{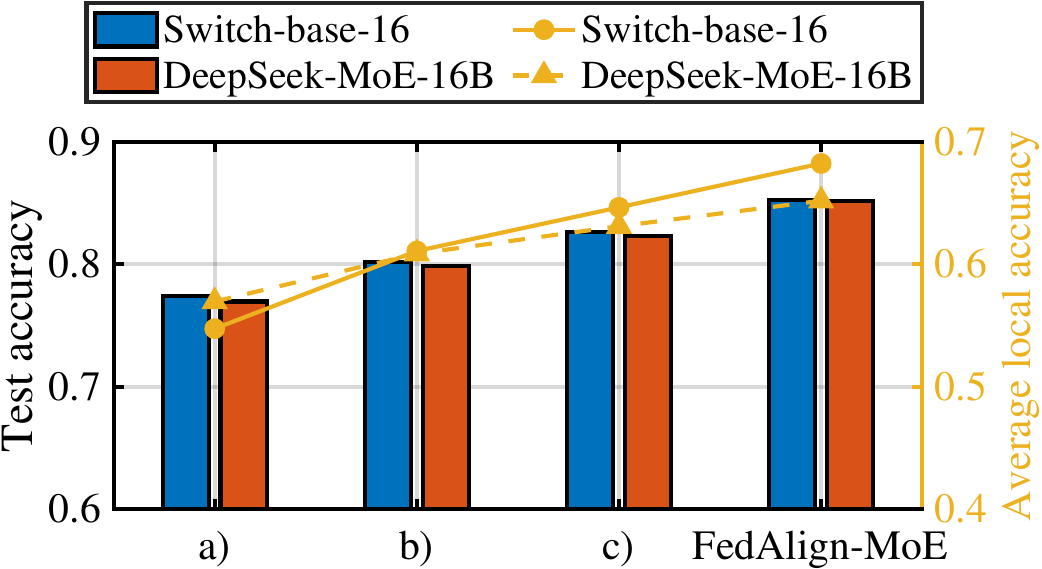}
}
\subfloat[MMLU]{
\includegraphics[width=0.49\linewidth]{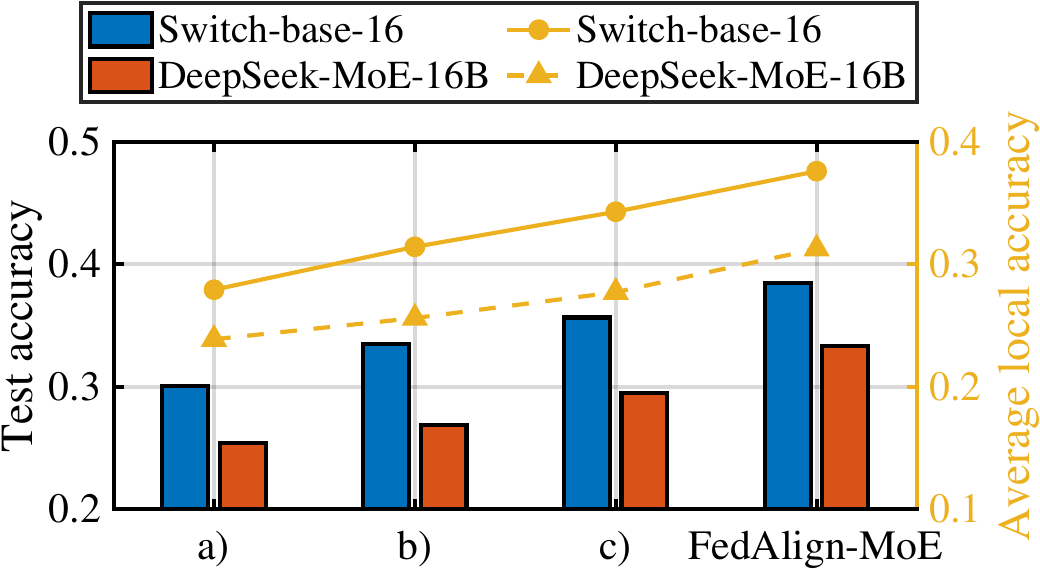}
}
\vspace{-1ex}
\caption{Consistency-based gating distribution alignment on AGNews and MMLU dataset using Switch-base-16 and DeepSeek-MoE-16B models.}
\label{fig:ablation_gating_alignment}
\vspace{-2ex}
\end{figure}

\vspace{-2ex}
\subsection{Ablation Study}
\subsubsection{\textbf{Consistency-based Gating Distribution Alignment}}
Fig.~\ref{fig:ablation_gating_alignment} reports the global test accuracy and the average local accuracy across clients on the Switch-base-16 and DeepSeek-MoE-16B models under non-IID data distributions. 
We compare the \name framework against three variants: a) standard parameter aggregation, which directly averages gating network parameters across clients; b) gating distribution alignment without routing consistency weighting, which aggregates client routing distributions uniformly; and c) gating distribution alignment without adaptive expert-level weights in client-side optimization.
The results show that standard parameter aggregation without routing distribution alignment leads to a substantial performance degradation in both local and global accuracy. 
Moreover, our consistency-based weighting results in a more robust global routing consensus, which improves global consistency while maintaining discriminative local gating preferences, yielding consistent gains in both local and global performance.
Adaptive expert-level regularization in  client-side optimization further enhances local accuracy as this soft constraint allows local gating networks to flexibly adapt to client-specific data characteristics, achieving stable optimization with strong cross-client generalization.

\subsubsection{\textbf{Semantic-Aware Expert Aggregation}}
\begin{figure}[t]
\vspace{-2ex}
\centering
\subfloat[Switch-base-16]{
\includegraphics[width=0.49\linewidth]{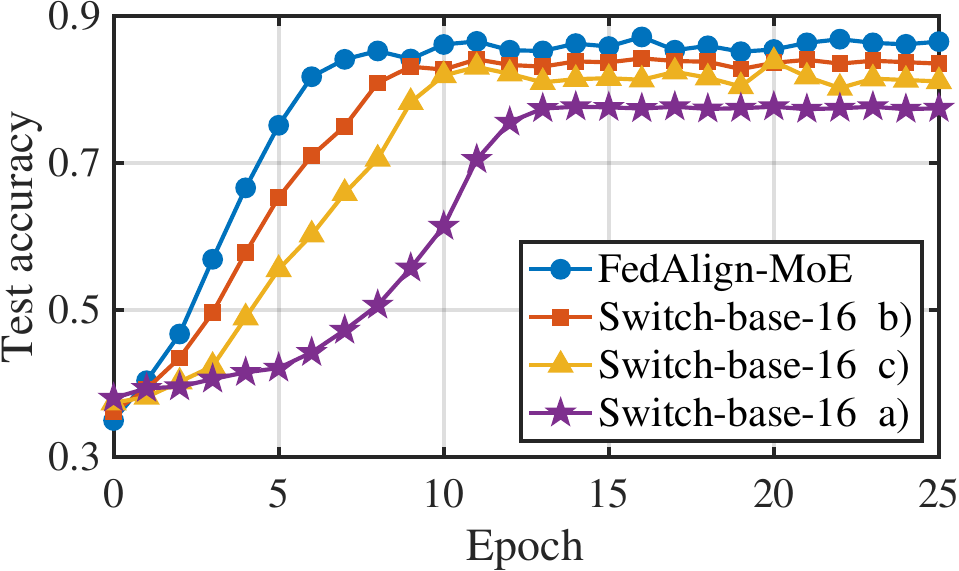}
}
\subfloat[DeepSeek-MoE-16B]{
\includegraphics[width=0.49\linewidth]{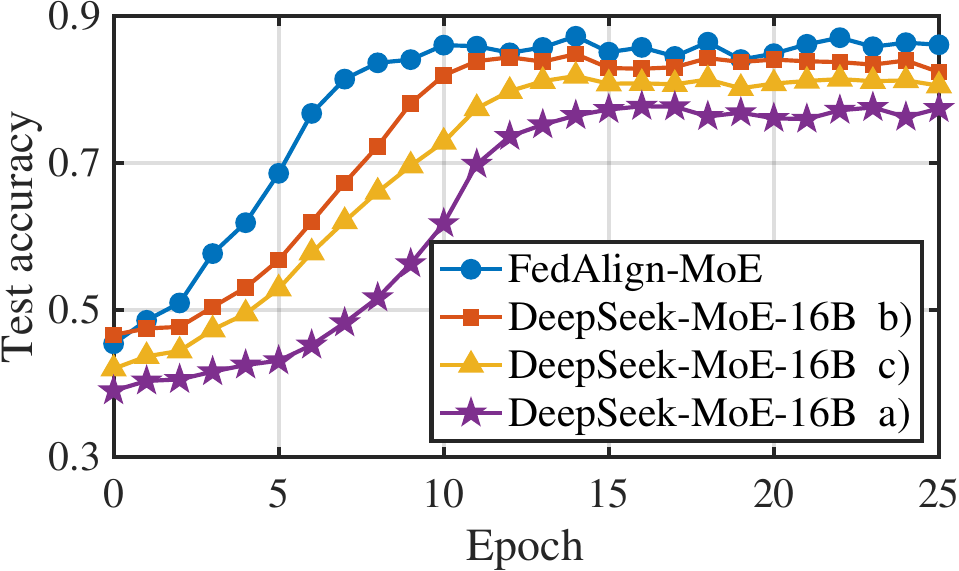}
} \\
\vspace{-2ex}
\subfloat[Switch-base-16]{
\includegraphics[width=0.49\linewidth]{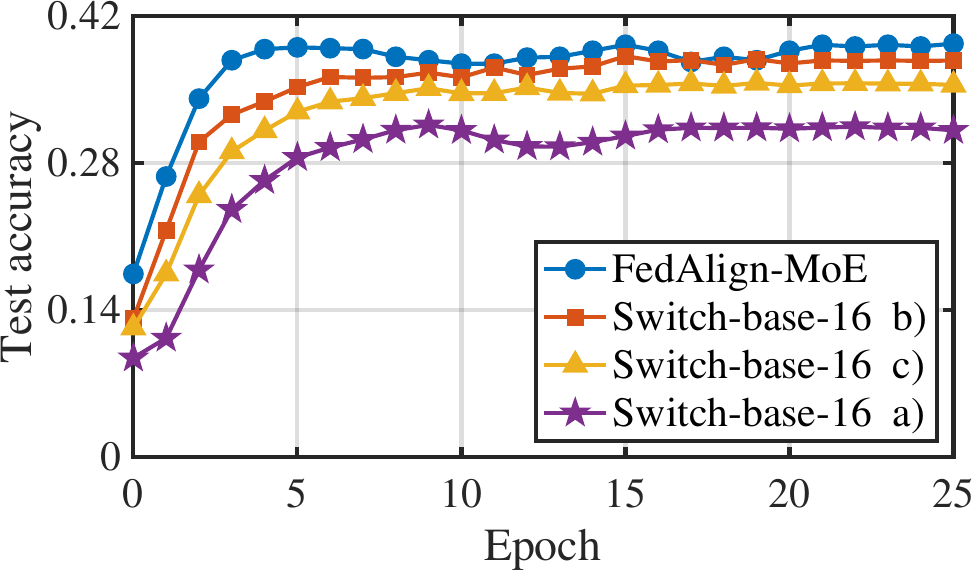}
}
\subfloat[DeepSeek-MoE-16B]{
\includegraphics[width=0.49\linewidth]{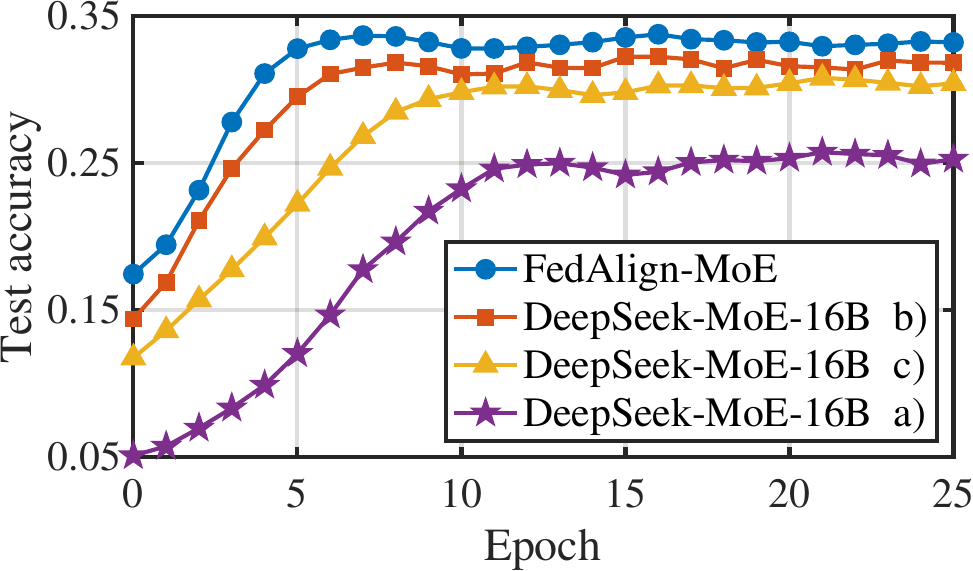}
}
\vspace{-1ex}
\caption{Semantic-aware expert aggregation on AGNews (Fig.~\ref{fig:ablation_expert_aggregation}(a)-(b)) and MMLU (Fig.~\ref{fig:ablation_expert_aggregation}(c)-(d)) datasets.}
\label{fig:ablation_expert_aggregation}
\vspace{-2ex}
\end{figure}

Fig.~\ref{fig:ablation_expert_aggregation} compares the expert aggregation performance of \name on the Switch-base-16 and DeepSeek-MoE-16B models under non-IID data distributions. 
We evaluate the test accuracy of expert aggregation under three variants: a) standard aggregation, which directly aggregates of expert parameters by index; b) region-conditioned aggregation without direction consensus, and c) region-conditioned aggregation without adaptive threshold, which utilizes a fixed global threshold.
Compared to standard  index-based aggregation, \name achieves consistently higher accuracy and faster convergence on both models, indicating that explicit semantic alignment via region-conditioned gating is crucial for effective expert aggregation.
When direction consensus is removed and expert semantic quantification only relies on input-space assignment, experts serving similar regions may still learn incompatible feature–label mappings, which introduces inconsistent optimization directions and degrades performance.
Moreover, the absence of adaptive thresholding undermines the selective filtering mechanism in aggregation, allowing misaligned updates to interfere with stable experts, thereby impairing expert specialization and slowing convergence.

\section{Related Work} \label{sec:related_work}
{\bf{MoE-based LLM Framework in Federated Learning:}}
By leveraging sparse activation mechanisms, MoE provides a promising paradigm for scaling LLM fine-tuning without sacrificing performance~\cite{dai2024deepseekmoe, fedus2022switch, zhan2024fedmoe, zhang2024mpmoe, xiao2025federated, xia2025fluxlog}, achieving superior performance in enhancing computation and communication efficiency.
Despite these advancements, data heterogeneity across client locations and user contexts introduces significant feature discrepancies for the integration of MoE within FL framework.
To address this, current investigations have extended MoE to personalized FL~\cite{guo2021pfl, zec2020specialized, feng2025pm, zhan2024fedmoe, yi2026pfedmoe, mei2024fedmoe}.
For instance, PFL-MoE~\cite{guo2021pfl} and pFedMoE~\cite{yi2026pfedmoe} alleviate heterogeneity through data-level personalization by combining the global aggregated model with locally fine-tuned MoE models. \needrev{More expert-centric approaches}, such as FedMoE-DA~\cite{zhan2024fedmoe} and PM-MoE~\cite{feng2025pm}, further identify and select critical experts that are most beneficial for client-specific data distributions, thereby enhancing adaptability to personalized local models on different clients.
However, these methods typically treat the MoE layer as a conventional weight collection and focus primarily on parameter-level aggregation by minimizing the distance between local and global weights, 
often overlooking the divergence in local routing preferences and the specialized expert roles driven by non-IID data, thus resulting in degraded performance due to the misalignment between the global model and local data characteristics.

{\bf{Gating Network Alignment during Federated MoE Aggregation:}}
Despite growing interest in integrating MoE models into FL, effectively aggregating local models with distinct client-specific preferences while preserving global performance poses significant challenges. 
Recent frameworks~\cite{li2020federated, karimireddy2020scaffold, zhang2023fedcp} stabilize gating behavior by employing distillation-based techniques or regularization terms to restrict routing divergence across clients. 
However, these solutions primarily \rev{enforce uniformity in expert selection} through a unified global gating, which often fails under data heterogeneity, as highly specialized local data lead to client-specific preferences that deviate significantly from the single global routing.
Alternatively, several works~\cite{zhan2024fedmoe, hu2025fft, xia2025fluxlog} maintain entirely private gating networks and aggregate only the activated experts to avoid cross-client behavior conflicts. 
However, without a mechanism to ensure global consistency in expert selection, this total localization leads to expert semantic mismatch where different clients may route identical semantic features to different expert indices, thereby destabilizing the functional specialization of experts during global aggregation and undermining the convergence of global MoE model.


{\bf{Expert Semantic Preservation for Federated MoE Fine-tuning: }}
Maintaining expert specialization is essential for MoE performance, leading to the development of the strategies for selective aggregation~\cite{fang2026hfedmoe, miao2025fedvla, mei2024fedmoe, radwan2025feddg} or cluster-based averaging~\cite{duan2021fedgroup, zhan2024fedmoe, isaksson2022adaptive, sievers2024advancing, sattler2020clustered} to support expert-level aggregation for better global model generalization and specialized knowledge preservation.
Selective aggregation frameworks like FedMoE~\cite{mei2024fedmoe} selectively incorporate expert updates only when they have significant contributions, preventing divergent or low-quality updates from polluting the global experts.
However, they typically judge an expert’s contribution within the parameter space rather than from its actual behavior in the functional space, ignoring the underlying semantic regions and the specific feature–label mappings an expert develops during local fine-tuning.
Client-level clustering methods, such as FedGroup~\cite{duan2021fedgroup} and FedMoE-DA~\cite{zhan2024fedmoe}, calculate similarity and group similar clients to improve global convergence.
While effective for homogeneous models, this approach suffers from a granularity mismatch in MoE architectures, as a single global client-level cluster cannot capture the divergent expert-level semantics that may belong to different functional groups within the same client. Such coarse-grained aggregation merges semantically mismatched updates across incompatible experts, eroding the structural advantages of the MoE-based LLMs and resulting in unstable convergence.

\section{Conclusion}\label{sec:conclusion}
In this paper, we have proposed a federated aggregation alignment framework, named \name, to address the routing preference divergence and expert semantic misalignment in MoE-based LLM fine-tuning under heterogeneous data distributions.
Our framework introduces two pivotal mechanisms: consistency-based gating distribution alignment and semantic-aware expert aggregation.
The consistency-based gating distribution alignment scheme aligns the distribution of gating outputs through consistency weighting, encouraging consistent cross-client gating behaviors while preserving the discriminative local preferences for personalized expert selection.
\needrev{Complementing this}, the semantic-aware expert aggregation strategy explicitly quantifies the semantic consistency of same-indexed experts across clients. Through a region-conditioned aggregation strategy and an adaptive thresholding mechanism, \name selectively aggregates expert updates from semantically aligned clients while suppressing conflicting ones, thereby ensuring that global experts maintain stable and specialized semantic roles.
Extensive experimental results have demonstrated that \name outperforms state-of-the-art benchmarks in both model accuracy and convergence stability, \jnlrev{demonstrating its practical deployability in mobile edge computing systems.}



\ifCLASSOPTIONcaptionsoff
  \newpage
\fi



%



\bibliographystyle{IEEEtran}
\bibliography{reference}

\end{document}